%% file: main.tex
\documentclass{article}

\usepackage[nonatbib, final]{neurips_2023}

\usepackage[square,numbers]{natbib} 
\usepackage[utf8]{inputenc} 
\usepackage[T1]{fontenc}    
\usepackage{hyperref}       
\usepackage{url}            
\usepackage{booktabs}       
\usepackage{amsfonts}       
\usepackage{nicefrac}       
\usepackage{microtype}      
\usepackage{xcolor}         

\title{Designing Long-term Group Fair Policies\\ in Dynamical Systems}

\input{format/usepackages}
\input{format/math-commands}

\input{format/paper-commands}

\author{%
  Miriam Rateike\\
  Saarland University\\
  Saarbrücken, Germany\\
 \And
  Isabel Valera \\
  Saarland University \\
  Max Planck Institute for Software Systems\\
  Saarbrücken, Germany  \\
 \AND
  Patrick Forré\\
    University of Amsterdam\\
  Amsterdam, The Netherlands  \\
  }

\begin{document}

\maketitle
\input{sections/00_Abstract}

\input{sections/01_Intro}
\input{sections/02_Example}
\input{sections/03_General-Principle}
\input{sections/04_Background}
\input{sections/05_Optimization-problem}
\input{sections/06_Long-term-targets}
\input{sections/07_Simulations}

\input{sections/08_Discussion}
\input{sections/09_Summary-Outlook}

\input{sections/acknowledgements}
\newpage

\bibliography{references}

\newpage

\cleardoublepage
\appendix

\input{sections/90_Appendix_Markov}

\input{sections/96_Appendix_clarifications}

\input{sections/91_Appendix_long-term-targets} 
\input{sections/92_Appendix_Simulation}

\input{sections/93_Appendix_Results-1}

\input{sections/93_Appendix_Results-2}

\input{sections/93_Appendix_Results-3}

\input{sections/93_Appendix_Results-4} 
\input{sections/93_Appendix_Results-5}

\input{sections/93_Appendix_Results-6}

\input{sections/94_Appendix_Guiding-example}

\end{document}

%% file: format/usepackages.tex
\usepackage{amsmath}
\usepackage{enumitem}
\usepackage{bbm}
\usepackage{tikz}
\usetikzlibrary{shapes,decorations,arrows,calc,arrows.meta,fit,positioning}
\tikzset{
    -Latex, auto, node distance = 0.5 cm and 0.5 cm, semithick,
    state/.style = {circle, draw, minimum width = 0.7 cm},
    const/.style = {minimum width = 0.7 cm},
    inter/.style = {rectangle, draw, minimum width = 0.7 cm, minimum height = 0.7 cm},
    point/.style = {circle, draw, inner sep = 0.04cm, fill, node contents = {}},
    bidirected/.style = {Latex-Latex,dashed},
    el/.style = {inner sep=2pt, align=left, sloped}
}

\usepackage{wrapfig}

\usepackage{caption}
\usepackage{subcaption}

\usepackage{caption}
\usepackage{subcaption}
\usepackage{amsthm}

\usepackage{hyperref}
\usepackage{url}

%% file: format/math-commands.tex
\newcommand{\N}{\mathbb{N}}
\newcommand{\R}{\mathbb{R}}

\newcommand{\Bcal}{\mathcal{B}}

\newcommand{\Hcal}{\mathcal{H}}
\newcommand{\Ical}{\mathcal{I}}

\newcommand{\Pcal}{\mathcal{P}}
\newcommand{\Qcal}{\mathcal{Q}}

\newcommand{\Scal}{\mathcal{S}}

\newcommand{\Ucal}{\mathcal{U}}

\newcommand{\Xcal}{\mathcal{X}}

\newcommand{\Zcal}{\mathcal{Z}}

\newcommand{\Xbf}{\mathbf{X}}

\newcommand{\dshto}{\dashrightarrow}

\newcommand{\E}{\mathbb{E}}

\DeclareMathOperator{\TV}{\mathrm{TV}}

\newcommand{\I}{\mathbbm{1}}

\newcommand{\lp}{\left ( }
\newcommand{\rp}{\right ) }

\newcommand{\lB}{\left [ }
\newcommand{\rB}{\right ] }
\newcommand{\lC}{\left \{ }
\newcommand{\rC}{\right \} }

\newcommand{\st}{\,\middle|\,}

\newtheorem{sa}{Theorem}[section]
\newtheorem{Thm}[sa]{Theorem}

\newtheorem{Cor}[sa]{Corollary}

\newtheorem{Def}[sa]{Definition}

\newtheorem{Not}[sa]{Notation}
\newtheorem{Asm}[sa]{Assumptions}

\newtheorem{Rem}[sa]{Remark}

%% file: format/paper-commands.tex
\usepackage{caption}

\everypar{\looseness=-1}
\newcommand{\inequity}{{\Ical}}

\newcommand{\Pbb}{\mathbb{P}}
\newcommand{\state}{{z}}
\newcommand{\State}{{Z}}
\newcommand{\chain}{{\left({\State_t}\right)_{t \in T}}}

\newcommand{\statefrom}{{z}}
\newcommand{\stateto}{{w}}
\newcommand{\statespace}{{\Zcal}}
\newcommand{\policy}{{\pi}}

\newcommand{\policyall}{{\pi(d \mid x, s)}}
\newcommand{\policyone}{{\pi(D=1 \mid x, s)}}
\newcommand{\policystar}{{\policy}^{\star}}

\newcommand{\labelfunctall}{{\ell(y \mid x, s)}}
\newcommand{\ylabelfunctall}{{f(x \mid y, s)}}

\newcommand{\labelfunctone}{{\ell(Y=1 \mid x, s)}}

\newcommand{\dynamicfunctall}{g(k \mid x, d, y, s)}
\newcommand{\ydynamicfunctall}{g(k \mid y, d, s)}

\newcommand{\kerneldef}{{P}}
\newcommand{\kerneldefpi}{{P_{\pi}}}
\newcommand{\kerneldefpis}{{P^s_{\pi}}}
\newcommand{\kernelpistar}{{T_{\pi^{\star}}}}

\newcommand{\kernel}{{T}}

\newcommand{\kernelspi}{{\kernel_{\pi}^s}}

\newcommand{\kernelpi}{{\kernel_{\policy}}}
\newcommand{\stationary}{{\mu}}
\newcommand{\stationaryt}{{\mu_t}}

\newcommand{\sensitiveall}{{\gamma(s)}}
\newcommand{\stationarypi}{{\stationary_{\policy}}}
\newcommand{\stationarypiall}{{\stationary_{\policy}(x \mid s)}}
\newcommand{\stationaryS}{{(\stationary^s)_{s\in \Scal}}}
\newcommand{\stationarys}{{\stationary^s}}
\newcommand{\stationarytall}{{\mu_t(x \mid s)}}
\newcommand{\ystationarytall}{{\mu_t(y \mid s)}}

\newcommand{\initialdistribution}{{\mu_0(x \mid s)}}
\newcommand{\yinitialdistribution}{{\mu_0(y \mid s)}}

\newcommand{\initialdist}{{\mu_0}}
\newcommand{\objective}{{J_{\text{LT}}}}

\newcommand{\Cconv}{{C_{\text{conv}}}}
\newcommand{\Clongterm}{{C_{\text{LT}}}}
\newcommand{\stationaryspi}{{\mu_{\pi}^s}}

\newcommand{\maxuitleopstar}{{\policy_{\texttt{EOP}}^{\star}}}
\newcommand{\maxqualstar}{{\policy_{\texttt{QUAL}}^{\star}}}

\newcommand{\eopunf}{{\texttt{EOPUnf}}}

\newcommand{\utility}{\Ucal}
\newcommand{\qualification}{{\Qcal}}

\newcommand{\irreducibility}{{\texttt{Irred}}}
\newcommand{\aperiodicity}{{\texttt{Aperiod}}}
\newcommand{\onesided}{{$\texttt{one-sided}$}}
\newcommand{\recourse}{{$\texttt{recourse}$}}
\newcommand{\discouraged}{{$\texttt{discouraged}$}}
\newcommand{\slow}{{$\texttt{slow}$}}
\newcommand{\medium}{{$\texttt{medium}$}}
\newcommand{\fast}{{$\texttt{fast}$}}

\newcommand{\randompol}{{\texttt{random}}}
\newcommand{\threspol}{{\texttt{threshold}}}
\newcommand{\biaspol}{{\texttt{bias}}}
\newcommand{\truepol}{{\texttt{true}}}

%% file: sections/00_Abstract.tex
\begin{abstract}
Neglecting the effect that decisions have on individuals (and thus, on the underlying data distribution) when designing algorithmic decision-making policies may increase inequalities and unfairness in the long term---even if fairness considerations were taken in the policy design process.
In this paper, we propose a novel 
framework for achieving long-term group fairness in dynamical systems, in which current decisions may affect an individual's features in the next step, and thus, future decisions. 
Specifically, our framework allows us to identify a time-independent policy that converges, if deployed, to the \emph{targeted} fair stationary state of the system in the long-term, independently of the initial data distribution.
We model the system dynamics with a time-homogeneous Markov chain and optimize the policy leveraging the Markov chain convergence theorem to ensure unique convergence.
We provide examples of different targeted fair states of the system, encompassing a range of long-term goals for society and policy makers.
Furthermore, we show how our approach facilitates the evaluation of different long-term targets by examining their impact on the group-conditional population distribution in the long term and how it evolves until convergence.
\end{abstract}

%% file: sections/01_Intro.tex
\section{Introduction}\label{sec:introduction}
\vspace{-5pt}

The majority of fairness notions that have been developed for trustworthy machine learning~\citep{hardt2016equality, dwork2012fairness}, assume an unchanging data generation process, i.e., a static system. Consequently, existing work has explored techniques to integrate these fairness considerations into the design of algorithms in static systems~\citep{hardt2016equality,dwork2012fairness,agarwal2018reductions,zafar2017bfairness, zafar2019fairness}.
However, these approaches neglect the dynamic interplay between algorithmic decisions and the individuals they impact, which have shown to be prevalent in practical settings~\citep{ chaney2018algorithmic, fuster2022predictably}. For instance, a decision to deny credit can lead to behavioral changes in individuals as they strive to improve their credit scores for future credit applications. This establishes a feedback loop from decisions to the data generation process, resulting in a shift in the data distribution over time, creating a dynamic system. 

Prior research has identified several scenarios where such dynamics can occur, including bureaucratic processes \citep{liu2018delayed}, social learning \citep{heidari2019effort}, recourse \citep{karimi2020algorithmic}, and strategic behavior \citep{hardt2016strategic, perdomo2020performative}.
Existing work on fair decision policies in dynamical systems has examined the effects of policies that aim to maintain existing static {group} fairness criteria in the short-term, i.e., in two-step scenarios~\citep{liu2018delayed,heidari2019effort} or over larger amount of time steps~\citep{zhang2020fair, creager2020causal, damour2020fairness}. These studies have demonstrated that enforcing static {group} fairness constraints in dynamical systems can lead to unfair data distributions and may perpetuate or even amplify biases~\citep{zhang2020fair, creager2020causal, damour2020fairness}.

Few previous work has attempted to meaningfully extend static fairness notions  to dynamic contexts by focusing on the long-term behavior of the system.
Existing approaches to learning long-term fair policies~\citep{perdomo2020performative, jabbari2017fairness, williams2019dynamic} assume unknown dynamics and
learn policies through iterative training within the reinforcement learning framework. 
While reinforcement learning offers flexibility and is, to some extent, model-agnostic, one of its major drawbacks lies in the requirement for large amounts of training data~\cite{henderson2018deep, dulac2021challenges, wang2016learning}, alongside the necessity for recurrent policy deployments over time.
 Successful applications of reinforcement learning typically occur in settings where a simulator or game is accessible~\cite{cutler2015real, osinski2020simulation}.
 However, in the real world, we can often not afford to satisfy such requirements.

To address these shortcomings, we propose to separate learning and estimation from decision-making and optimization.
We start with a modeling approach of the main relevant (causal) mechanisms of the real world first and 
require access to a sufficient amount of data to reliably estimate these. 
The main contribution of this paper then lies in proposing a method of how to use this information to find a policy that leads to a stable long-term fair outcome as an equilibrium state.

We introduce a principle that can be applied to various (causal) models to learn policies aimed at achieving long-term group fairness, along with a computational optimization approach to solve it. 
Our framework can be thought of as a three-step process: Given sufficient data to estimate (causal) mechanisms, we i) define the characteristics of a long-term fair distribution in the decision-making context; ii) transform this definition into a constrained optimization problem; iii) which we then solve.
Importantly, existing long-term group fairness targets~\citep{chi2022towards, wen2021algorithms, yin2023long, yu2022policy} can be formulated as such long-term fair distribution.

Inspired by previous work~\citep{zhang2020fair}, we adopt Markov chains as a framework to model system dynamics.
We propose an \emph{optimization problem} to find a policy that, if found, guarantees that the system converges, irrespective of the initial state, to the pre-defined targeted fair, stationary data distribution.
Such policy offers consistency in decision-making, enhancing stakeholder trust and predictability of decision processes.
Furthermore, the policy is guaranteed to converge from any 
starting distribution, which makes it robust to covariate shift.

Our work differs from research on fair sequential decision learning under feedback loops, where decisions made at one time step influence the training data observed at the subsequent step \citep{kilbertus2019fair, Rateike2022DontTI, bechavod2019equal, joseph2016fairCB}.
In this scenario, decisions introduce a sampling bias, but do not affect the underlying generative process, as in our case.
In our case, decisions influence the underlying data-generating process and consequently shift the data distribution. 
Our work also diverges from research focused on developing robust machine learning models that can perform well under distribution shifts, where deployment environments may differ from the training data environment~\citep{quinonero2008dataset}. 
Unlike the line of research that considers various sources of shift~\citep{makar2022fairness, adragna2020fairness, schrouff2022maintaining}, our approach leverages policy-induced data shifts to guide the system towards a state that aligns with our defined long-term fairness objectives. 
Rather than viewing data shifts as obstacles to overcome, we utilize them as a means to achieve fairness goals in the long term.

While our framework can be applied to various dynamical systems, we first provide a guiding example (\S~\ref{sec:example}).
We then provide a framework for policy makers to design fair policies that strategically use system dynamics to achieve effective fair algorithmic decision-making in the long term (\S~\ref{sec:general-goal}) together with a general optimization problem that allows solving it computationally (\S~\ref{sec:general-principle}).
We then exemplify targeted fair states for the system, leveraging {existing fairness} criteria (\S~\ref{sec:goals}).
Following previous work~\citep{creager2020causal, damour2020fairness}, we use simulations to systematically explore the convergence and behavior of different long-term policies found by our framework (\S~\ref{sec:simulations}).
We conclude with a discussion (\S~\ref{sec:discussion}), followed by a summary and outlook (\S~\ref{sec:summary-outlook}).

%% file: sections/02_Example.tex
\vspace{-10pt}
\section{Guiding Example}\label{sec:example}
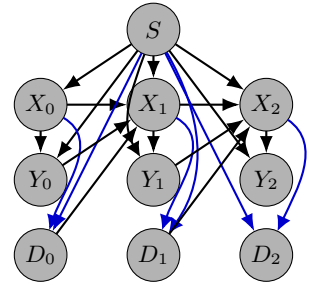
\begin{wrapfigure}{r}{0.3\textwidth}
    \centering
    \scalebox{1}{
        \input{figures/data-gen-x-y}
    }
    \caption{Data generative model.
    Time steps (subscript) ${t=\{0, 1, 2\}}$. Policy $\pi$ blue.}
    \label{fig:data-gen-x-y}
        \vspace{-7pt}
\end{wrapfigure}

\vspace{-5pt}
%
We present a guiding example. {Note, however, that our framework can also be applied framework to other generative processes (see Appendix~\ref{apx:example}).}
We assume a data generative model for a credit lending scenario~\citep{ liu2018delayed, creager2020causal,damour2020fairness} (see Figure~\ref{fig:data-gen-x-y}).
%
%
\paragraph{Data generative model.} Let an individual with protected attribute $S$ (e.g. gender) at time $t$ be described by a non-sensitive feature $X_t$ (e.g. credit score as a summary of monetary assets and credit history) and an outcome of interest $Y_t$ 
(e.g. repayment ability).
We assume the sensitive attribute 

to remain immutable over time, and drop the attributes time subscript.
For simplicity, we assume binary sensitive attribute and outcome of interest ${S, Y \in \{0,1\}}$ and a one-dimensional discrete non-sensitive feature ${X \in \mathbb{Z}}$.
Let the population's sensitive attribute be distributed as ${\sensitiveall:=\Pbb(S\!=\!s)}$ and remain constant over time. 
We assume $X$ to depend on $S$, such that the group-conditional feature distribution at time $t$ is ${\stationarytall:=\Pbb(X_t\!=\!x \mid S\!=\!s)}$.
For example, different demographic groups may have different credit score distributions due to structural discrimination in society. 
The outcome of interest is assumed to depend on $X$ and (potentially) on $S$ resulting in the label distribution ${\labelfunctall:= \Pbb(Y_t\!=\!y \mid X_t\!=\!x, S\!=\!s)}$. 
For example, payback probability may be tied to factors like income, which can be assumed to be encompassed within a credit score.
We assume that there exists a policy that takes binary loan decisions 
based on $X$ and (potentially) $S$
and decides with probability $\policyall:=\Pbb(D_t\!=\!d\mid X_t\!=\!x, S\!=\!s)$. 
Consider dynamics where a decision $D_t$ at time step $t$ directly influences an individual's features $X_{t+1}$ at the next step. We assume, the transition from the current feature state $X_t$ to the next state $X_{t+1}$ depends additionally on the current features, outcome $Y_t$, and (possibly) the sensitive attribute $S$.
For example, after a positive lending decision, an individual's credit score may rise due to successful loan repayment, with the extent of increase (potentially) influenced by their sensitive attribute.
Let the probability of an individual with $S\!=\!s$ transitioning from a credit score of $X_t\!=\!x$ to $X_{t+1}\!=\!k$ in the next step, denoted as the dynamics $\dynamicfunctall:=\Pbb(X_{t+1}=k | X_t\!=\!x, D_t\!=\!d, Y_t\!=\!y, S\!=\!s)$
Importantly, the next step feature state depends only on the present feature state, and not on any past states. 
%
\paragraph{Dynamical System.} 
We can now describe the evolution of the group-conditional feature distribution $\stationarytall$ over time $t$. 
The probability of a feature change from $X_t = x$ to $X_{t+1} = k$ in the next step given $S=s$ is obtained by marginalizing out $D_t$ and $Y_t$, resulting in 
\begin{align}\label{eq:example-x-y-kernel}
\begin{split}
        & \Pbb(X_{t+1}\!=\!k \mid X_{t}\!=\!x, S\!=\!s) = \sum_{d, y} \dynamicfunctall \policyall \labelfunctall.
\end{split}
\end{align}
%
These transition probabilities together with the initial distribution over states $\initialdistribution$ define the behavior of the dynamical system.
In our model, we assume time-independent dynamics $\dynamicfunctall$, where feature changes in response to decisions and individual attributes remain constant over time (e.g., through a fixed bureaucratic policy determining credit score changes based on repayment behavior).
We also assume that the distribution of the outcome of interest conditioned on an individual's features $\labelfunctall$ remains constant over time (e.g., individuals need certain assets, summarized in a credit score, to repay).
Additionally, we assume that the policy $\policyall$ can be chosen by a policy maker and may depend on time. Under these assumptions, the probability of a feature change depends solely on policy $\policy$ and sensitive feature $S$.

\paragraph{Targeted Fair Distribution.}
Consider a bank using policy $\policy$ for loan approvals. 
While maximizing total profit, the bank also strives for fairness by achieving equal credit score distribution across groups~\citep{damour2020fairness}.
This means, at time $t$ the probability of having a credit score $x$ should be equal for both sensitive groups:
$\stationary_t(x \mid S=0) = \stationary_t(x \mid S=1)$ 
for all $x \in \Xcal$.
If credit scores are equally distributed, the policy maker aims to preserve this equal distribution in the next time step: 
\begin{align}\label{eq:example-x-y-stationary}
\begin{split}
     & \stationary_{t+1}(k \mid s) = \sum_{x} \stationary_{t}(x \mid s) \Pbb(X_{t+1}=k \mid X_t =x, S=s) 
\end{split}
\end{align}
for all $k \in \Xcal$, $s \in \{0,1\}$.
This means, the credit score distribution remains unchanged (stationary) when multiplied by the transition probabilities defined above.
The policy maker's task is then to find a policy $\policy$ that guarantees the credit score distribution to converge to the targeted fair distribution.

%% file: figures/data-gen-x-y.tex
  \begin{tikzpicture}[every node/.style={inner sep=0,outer sep=0, font=\small}, scale=0.7]
        \node[state, fill=gray!60] (x0) at (0,0) {{$X_0$}};
        \node[state, fill=gray!60] (x1) [right  =  0.8 cm of x0]  {$X_1$};
        \node[state, fill=gray!60] (x2) [right  =  0.8 cm of x1]  {$X_2$};
        \node[state, fill=gray!60] (s) [above  =  0.3cm of x1]  {$S$};
         \node[state, fill=gray!60] (y0) [below  =  0.3 cm of x0] {$Y_0$};
        \node[state, fill=gray!60] (y1) [below  =  0.3 cm of x1]  {$Y_1$};
        \node[state, fill=gray!60] (y2) [below  =  0.3 cm of x2]  {$Y_2$};
        \node[state,  fill=gray!60] (d0) [below  =  0.3 cm of y0] {$D_0$};
        \node[state,  fill=gray!60] (d1) [below  =  0.3 cm of y1]  {$D_1$};
        \node[state,  fill=gray!60] (d2) [below  =  0.3 cm of y2]  {$D_2$};
        \path (x0) edge [thick](x1);
        \path (x1) edge [thick] (x2);      
        \path (s) edge [thick](y0);
        \draw [thick, -{Latex[length=2mm,width=2mm]}] (s) to [out=250,in=120] (y1);
        \path (s) edge [thick](y2);
        \path (s) edge [thick, black!20!blue](d0);
        \draw [thick, -{Latex[length=2mm,width=2mm]}, black!20!blue] (s) to [out=300,in=55] (d1);
        \path (s) edge [thick, black!20!blue](d2);
        \path (s) edge [thick](x0);
        \path (s) edge [thick](x1);
        \path (s) edge [thick](x2);
        \path (x0) edge [thick](y0);
        \path (x1) edge [thick](y1);
        \path (x2) edge [thick](y2);
        \draw [thick, -{Latex[length=2mm,width=2mm]}, black!20!blue] (x0) to [out=330,in=71] (d0);
        \draw [thick, -{Latex[length=2mm,width=2mm]}, black!20!blue] (x1) to [out=330,in=71] (d1);
        \draw [thick, -{Latex[length=2mm,width=2mm]}, black!20!blue] (x2) to [out=325,in=65] (d2);
        
        \path (d0) edge [thick](x1);
        \path (d1) edge [thick](x2);
        \path (y0) edge [thick](x1);
        \path (y1) edge [thick](x2);
\end{tikzpicture}

%% file: sections/03_General-Principle.tex
\section{{Designing Long-term Fair Policies}}\label{sec:general-goal}
\vspace{-5pt}
After having introduced the guiding example, 
we now move to a more general setting of time-homogeneous Markov chains that depend on a policy and sensitive features.
%
\subsection{Background: Time-homogeneous Markov Chains}
\vspace{-5pt}
We remind the reader of the formal definition of time-homogeneous Markov chains with discrete states space and draw on the following literature for definitions~\citep{freedman2017convergence}. For a formulation for general state spaces refer to the Appendix~\ref{apx:markov} or~\citep{meyn2012markov}.
\begin{Def}[Time-homogeneous Markov Chain]
A time-homogeneous Markov chain on a discrete space $\statespace$ with transition probability $\kerneldef$ is a sequence of random variables $\left({\State_t}\right)_{t \in T}$ with joint distribution $\Pbb$, such that for every $t \in T$ and $\statefrom, \stateto \in \statespace$ we have $\Pbb(\State_{t+1}\!=\!\stateto \mid \State_t\!=\!\statefrom) =\kerneldef(\statefrom, \stateto)$.
\end{Def}
In a Markov chain, each event's probability depends solely on the previous state. Recall that the transition probabilities must satisfy $\kerneldef(\statefrom, \stateto) \geq 0$ for all $\statefrom, \stateto$, and $\sum_{\stateto} \kerneldef(\statefrom, \stateto) = 1$ for all $\statefrom$. The guiding example can be seen as a Markov chain with state space $\Xcal$ and transition probabilities (\ref{eq:example-x-y-kernel}). We have stated that the policy maker aims to achieve a fair stationary distribution (\ref{eq:example-x-y-stationary}). To formally define this, we introduce the following concept:

\begin{Def}[Stationary Distribution] \label{def:stationary}
A stationary distribution of a time-homogeneous Markov chain $(\statespace, \kerneldef)$ 
is a probability distribution $\stationary$, such that
$\stationary = \stationary \kerneldef$. 
More explicitly, for every $\stateto \in \statespace$ the following needs to hold: $\stationary(\stateto) = \sum_{\statefrom} \stationary(\statefrom) \cdot \kerneldef(\statefrom,\stateto)$.
\end{Def}
%
In words, the distribution $\stationary$ remains unchanged when multiplied by the transition kernel $\kerneldef$. 
%
\subsection{\
The Objective for Long-term Fair Policies}
\vspace{-5pt}
We generalize the provided example to time-homogeneous Markov chains that depend on a policy $\policy$ and a sensitive attribute $S$. The population's feature distribution over time is represented by a time-homogeneous Markov chain $\chain$ with a general state space $\statespace$. The transition probabilities that depend on the sensitive attribute $S$ and policy $\policy$ are captured by the transition probabilities $\kerneldefpis$. Suppose a policy maker aims to achieve a fair distribution $\stationaryS$.
The goal for the policy maker is then to find a distribution $\stationaryS$ and policy $\policy$ such that the induced kernel $\kerneldefpis$ converges to the distribution $\stationaryS$, and the distribution $\stationaryS$ satisfies the defined fairness constraints.

Now, consider a scenario where our society is already in a fair state $\stationaryS$. In this case, the policy maker would aim to find policy $\policy$ that defines a transition probability $\kerneldefpis$ such that the next state remains fair. More formally, we would seek to satisfy the following equation:
\begin{align}\label{eq:general-objective}
    \stationarys = \stationarys \kerneldefpis
\end{align}
for all $s \in \Scal$.
This can be seen as a generalization of (\ref{eq:example-x-y-stationary}). Therefore, the fair distribution $\stationaryS$ should be the stationary distribution of the Markov chain defined by $(\statespace, \kerneldefpis)$. Any policy that aims for the fair stationary state $\stationaryS$ will eventually need to find a policy that satisfies (\ref{eq:general-objective}) to at least transition from a fair state to a fair state in the long term. In this sense (\ref{eq:general-objective}) defines the fundamental problem of finding long-term fair policies in these settings.
To find a policy that ensures convergence to the desired fair distribution, we present a general optimization problem in \S~\ref{sec:general-principle}. This utilizes the Markov Convergence Theorem, which we discuss next.

%% file: sections/04_Background.tex
\section{Background on Markov Chain Convergence Theorem}\label{sec:background}
\vspace{-5pt}
The Markov Convergence Theorem establishes conditions for a time-homogeneous Markov chain to converge to a unique stationary distribution, regardless of the initial distribution. In our model, the transition probabilities depend on the sensitive attribute, and we will apply in (\ref{op:general}) the Markov Convergence theorem separately to each group's transition probabilities. We thus drop the superscript $s$.
%
\begin{Thm}[Markov Convergence Theorem]\label{thm:markov-convergence} 
Let $\chain$ 
be an irreducible and aperiodic time-homogeneous Markov chain with discrete state space $\statespace$ and transition matrix $\kerneldef$. Then the marginal distribution $\Pbb(\State_t)$ converges to the unique stationary distribution $\stationary$ as $t$ approaches infinity (in total variation norm), regardless of the initial distribution $\Pbb(\State_0)$.
\end{Thm}
%
In words, the Markov Convergence Theorem states that, regardless of the initial distribution, the state distribution of an irreducible and aperiodic Markov chain eventually converges to the \emph{unique} stationary distribution. We now provide definitions for irreducibility and aperiodicity.

\begin{Def}[Irreducibility]\label{def:irreducibility}
A time-homogeneous Markov chain is considered irreducible if, for any two states $\statefrom, \stateto \in \statespace$, there exists a 
$t> 0$ such that $\kerneldef^t(\statefrom, \stateto) > 0$, where ${\kerneldef^t(\statefrom, \stateto)} = {\Pbb(\State_{t}=\stateto \mid \State_0=\statefrom)}$ represents the probability of going from $\statefrom$ to $\stateto$ in $t$ steps. 
\end{Def}
%
In other words, irreducibility ensures that there is a positive probability of reaching any state $\stateto$ from any state $\statefrom$ after some finite number of steps. Note, for discrete state space $\statespace$, every irreducible time-homogeneous Markov chain has a unique stationary distribution~(Thm. 3.3~\citep{freedman2017convergence}).
%
\begin{Def}[Aperiodicity]\label{def:aperiodicity}
Consider an irreducible time-homogeneous Markov chain $(\statespace, \kerneldef)$.
Let $R(\state)=\left\{t \geq 1: \kerneldef^t(\state, \state)>0\right\}$ be the set of return times from $\state \in \statespace$, where $\kerneldef^t(\state, \state)$ represents the probability of returning to state $\state$ after $t$ steps. The Markov chain is aperiodic if and only if the greatest common divisor (gcd) of $R(\state)$ is equal to 1:
$gcd(R(\state))=1$ for all $\state$ in $\statespace.$ 
\end{Def}
%
In words, aperiodicity refers to the absence of regular patterns in the sequence of return times to state $\state$, i.e., the chain does not exhibit predictable cycles or periodic behavior. 
For general state spaces the Markov Convergence Theorem can be proven under Harris recurrence, aperiodicity and the existence of a stationary distribution~\citep{meyn2012markov} (see Apx.~\ref{apx:markov}).

%% file: sections/05_Optimization-problem.tex
\section{
The Optimization Problem}\label{sec:general-principle}
\vspace{-5pt}
We now reformulate objective (\ref{eq:general-objective}) into a computationally solvable optimization problem for finding a time-independent policy. This policy, if deployed, leads the system to convergence to a fair stationary state in the long term, regardless of the initial data distribution.

\begin{Def}[General Optimization Problem]
Assume a time-homogeneous Markov chain $(\statespace, \kerneldefpi)$ defined by state space $\statespace$ and kernel $\kerneldefpis$.
To find policy $\policy$ that ensures the Markov Chain's convergence to a unique stationary distribution $\stationaryS$, while minimizing a
fair long-term objective $\objective$ and adhering to a set of 
fair long-term constraints $\Clongterm$, we propose the following optimization problem:
%
\begin{align}\label{op:general}
    \begin{split}
        &\min_{\pi} \quad \objective (\stationaryS, \policy)  \quad \quad \text{subj. to} \quad \Clongterm (\stationaryS, \policy)  \geq 0 ; 
         \quad  \Cconv(\kerneldefpis) \geq 0 \, \forall s\\
    \end{split}
\end{align}
where $\Cconv$ are convergence criteria according to the Markov Convergence Theorem. 
\end{Def}

In words, we aim to find a policy $\policy$ that minimizes a long-term objective $\objective$ subject to long-term constraints $\Clongterm$ and convergence constraints $\Cconv$. 
The objective $\objective$ and constraints $\Clongterm$ are dependent on the policy-induced stationary distribution $\stationaryS$, which represents the long-term equilibrium state of the data distribution and may also depend directly on the policy $\policy$. 
In \S~\ref{sec:goals}, we provide various instantiations of long-term objectives and constraints to illustrate different ways of parameterizing them.
Convergence constraints $\Cconv$ are placed on the kernel $\kerneldefpis$ and guarantee convergence of the chain to a \emph{unique stationary distribution for any starting distribution} according to the Markov Convergence Theorem (Def.\ref{thm:markov-convergence}). 
The specific form of $\Cconv$ depends on the properties of the Markov chain, such as whether the state space is finite or continuous. 
In the following, we refer to the notation $\stationarypiall$ when we are interested in $\stationaryS$ at certain values $x$ and $s$.
%
\paragraph{Solving the Optimization Problem.}
In our example, the Markov chain is defined over a categorical feature $X$ (credit score), resulting in a finite state space. In this case, the optimization problem becomes a linear constrained optimization problem and we can employ any efficient black-box optimization methods for this class of problems (e.g.,~\cite{kraft1988software}). We detail this for our example:
The convergence constraints $\Cconv$ are determined by the aperiodicity and irreducibility properties of the corresponding Markov kernel (see \S~\ref{sec:background}). 
A sufficient condition for irreducibility is $\irreducibility(\pi):= \sum_{i = 1}^{n} \left(\kernelspi\right)^n \geq \mathbf{0} \, \forall s$, where
$n$ is the number of states ($n = |X|$), and $\mathbf{0}$ denotes the matrix with all entries equal to zero. 
A sufficient condition for aperiodicity requires that the diagonal elements of the Markov kernel are greater than zero: $\aperiodicity(\policy): = \kernelspi(x, x) > 0 \, \forall x, s$. 
The group-dependent stationary distribution $\stationaryspi$ based on $\kernelspi$ can be computed via eigendecomposition~\citep{weber2017eigenvalues}.
In the next section we introduce various objective functions $\objective$ and constraints $\Clongterm$ that capture notions of profit, distributional, and predictive fairness. Importantly, for finite state spaces, these objectives and constraints are linear.
While our general optimization problem remains applicable in the context of an infinite state space, solving it becomes more challenging due to the potential introduction of non-linearities and non-convexities.

%

%% file: sections/06_Long-term-targets.tex
\section{Targeted Fair States}\label{sec:goals}
\vspace{-5pt}
Our framework enables users to define their preferred long-term group fairness criteria. 
Here, we present examples of how long-term fair targets can be quantified by defining a long-term objective $\objective$ and long-term constraints $\Clongterm$ in (\ref{op:general}). 
We provide these examples assuming discrete $X$ and binary $D, Y, S$ as in our guiding example (\S~\ref{sec:example}). Note, our framework allows enforcing common long-term fairness and reward notions (see Appendix~\ref{apx:clarification-definition-long-term-fairness}).
%
\subsection{Profit}\label{sec:goals-profit}
\vspace{-5pt}
Assume that when a granted loan is repaid, the bank gains a profit of $(1\!-\!c)$; when a granted loan is not repaid, the bank faces a loss of $c$; and when no credit is granted, neither profit nor loss occurs.
We quantify this profit as utility~\citep{kilbertus2020fair, corbett2017algorithmic}, considering a cost associated with positive decisions denoted by $c \in [0,1]$, in the following manner: $\utility(\policy; c)= \sum_{x, s}\policyone\left(\labelfunctone -c\right) \stationarypiall \sensitiveall$,
where $\policyone$ is the probability of a positive policy decision, $\labelfunctall$ the positive ground truth distribution, $\stationarypiall$ the stationary group-dependent feature distribution, and $\sensitiveall$ the distribution of the sensitive feature.

A bank's objective may be to maximize utility (minimize financial loss, i.e., $\objective:= - \utility(\policy, c)$). In contrast, a non-profit organization may aim to constrain its policy by maintaining a minimum profit level $\epsilon \geq 0$ over the long term to ensure program sustainability ($\Clongterm:=\utility(\policy; c) - \epsilon$).
%
\subsection{Distributional Fairness}\label{sec:goals-distributional-fairness}
\vspace{-5pt}
Policy makers may also be interested in specific characteristics of a population's features $X$ or qualifications $Y$ (ground truth) on a group level.
We measure group qualification $\qualification$ as the group-conditioned proportion of positive labels assigned to individuals~\citep{zhang2020fair} as 
${\qualification^s(\policy \mid s)} = \sum_{x} 
     \labelfunctone \stationarypiall$,
where $\labelfunctone$ is the positive ground truth distribution, and $\stationarypiall$ describes the stationary group-dependent feature distribution. We measure inequity (of qualifications) as ${\inequity:=\mid \qualification(\policy \mid S=0) - \qualification(\policy \mid S=1) \mid}$.

To promote financial stability, a policy maker like the government may pursue two different objectives. Firstly, they may aim to minimize default rates using the objective $\objective:={ -\sum_s\qualification(\policy \mid s) \sensitiveall}$. Alternatively, if the policy maker intends to increase credit opportunities, they may seek to maximize the population's average credit score with the objective ${\objective:={ -\sum_s \frac{1}{|X|}\sum_x\stationarypiall \sensitiveall}}$, where $|X|$ represents the state space size.
To achieve more equitable credit score distributions, the policy maker could impose the constraint $\Clongterm:={\epsilon - \mid \stationarypi(x \mid S=0) - \stationarypi(x\mid S= 1) \mid  \forall x }$. However, depending on the generative model, this approach might not eliminate inequality in repayment probabilities.
In such cases, the policy maker may aim to ensure that individuals have the same payback ability using the constraint $\Clongterm:=\epsilon - \inequity$. Note that measuring differences in continuous or high-dimensional distributions requires advanced distance measures.
Additionally, prioritizing egalitarian distributions may not always align with societal preferences~\citep{barsotti2022minmax, martinez2020minimax} (see  Appendix~\ref{apx:goals}).
Finally, equal credit score distributions or repayment probabilities may not guarantee equal access to credit, we thus next introduce predictive group fairness measures.

\subsection{Predictive Fairness}\label{sec:goals-predictive-fairness}
\vspace{-5pt}
Ensuring long-term predictive fairness can help a policy maker meet regulatory requirements and maintain public trust. One example of a predictive group unfairness measure is \emph{equal opportunity}~\citep{hardt2016equality}: ${\eopunf(\policy)}\!=\mid\!\Pbb_{\policy}(D\!=\!1\! \mid\!Y\!=\!1, S\!=\!0){-\Pbb_{\policy}(D\!=\!1\!\mid\!Y\!=\!1, S\!=\!1)\!\mid}\!$.  
This measures the disparity in the chance of loan approval for eligible loan applicants based on their demographic characteristics. Note: ${\Pbb_{\policy}(D\!=\!1\!\mid\!Y\!=\!1, S\!=\!s)=\frac{\sum_x\policyone \labelfunctone \stationarypiall}{\sum_x\labelfunctone \stationarypiall}}$. 

In the fairness literature, it is common for a policy maker to define a maximum tolerable unfairness threshold as $\epsilon \geq 0$, expressed as $\Clongterm:= \epsilon - \eopunf$.
Alternatively, they may aim to minimize predictive unfairness $\eopunf$ over the long term by imposing $\objective:=\eopunf(\policy)$.
Note, our framework also allows for other group fairness criteria, such as demographic parity~\citep{dwork2012fairness} or sufficiency~\citep{chouldechova2017fair}.\\

In this section, we presented various long-term goals as illustrative examples for lending policies. For methods to impose constraints on the types of policies under consideration, please refer to Appendix~\ref{apx:goals}. This section serves as a starting point for discussions on these objectives and we encourage the exploration of a wider range of long-term targets by drawing inspiration from existing research in social sciences and economics, while also involving affected communities in defining these objectives. In the following section, we demonstrate how our approach enhances the understanding of the interplay between diverse long-term goals and constraints.

%% file: sections/07_Simulations.tex
\section{Simulations}\label{sec:simulations}
\vspace{-5pt}
We validate
our proposed optimization problem formulation in semi-synthetic simulations.
Using our guiding example with real-world data and assumed dynamics, we first demonstrate that the policy solution, if found, converges to the targeted stationary state (\S~\ref{sec:sim_validation}). Then, we demonstrate how our approach helps to analyze the interplay between long-term targets and dynamics (\S~\ref{sec:sim_dynamics}).
For additional results see Appendix~\ref{apx:results}. 

\paragraph{Data and General Procedure.} 
We use
the real-world FICO loan repayment dataset~\citep{fico2007}, with data pre-processing from~\citep{fairmlbook}. 
It includes a one-dimensional credit score $X$, which we discretize into four bins for simplicity, and a sensitive attribute $S$ that we binarize: Caucasian ($S=1$) and African American ($S=0$). 
From this dataset, we estimate the initial feature distribution $\initialdistribution$, label distributions $\labelfunctall$, and sensitive group ratios $\sensitiveall$. Note, the FICO dataset provides probability estimates.
For results under estimated probabilities and dynamics when labels are partially observed, refer to the Appendix~\ref{apx:additional-estimations}.
Since FICO is a static dataset, 
we assume dynamics $\dynamicfunctall$.
We first apply the general principle (\ref{op:general}) to formulate an optimization problem 
via long-term objectives $\objective$ and long-term constraints $\Clongterm$ and convergence constraints $\Cconv$. 
Next, we solve the optimization problem. 
Using the found policy $\policystar$ and the resulting Markov kernel $\kernelpistar$, we generate the feature distribution across 200 steps.
See Appendix~\ref{apx:sim} for details.
%

We solve the problem using the Sequential Least Squares Programming method from scikit-learn~\citep{pedregosa2011scikit}, initializing it (warm start) with a uniform policy where all decisions are random ($\policyone=0.5$ $\forall s,x$).
See Appendix~\ref{apx:sim} for details.
\begin{figure}
    \hfill
    \begin{subfigure}{0.49\textwidth}
        \centering
        \includegraphics[width=\textwidth]{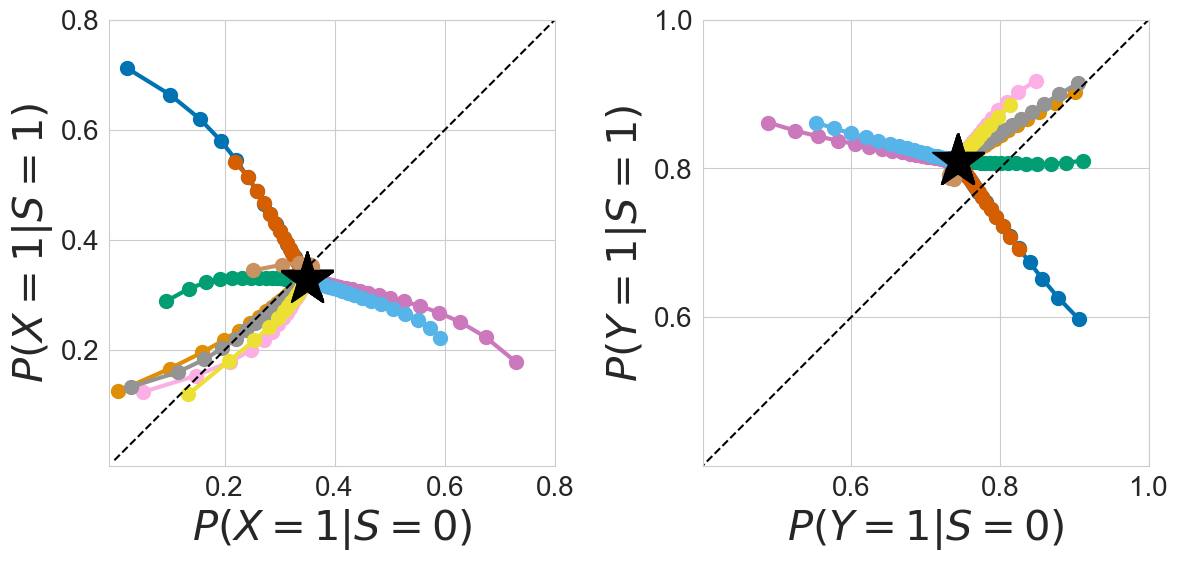}
        \caption{Convergence: $\maxuitleopstar$ to unique stationary distribution $\star$. $200$ time steps. Colors: 10 random initial feature distributions. Feature $X\!=\!1$ left, outcome $Y$ right. Equal distribution dashed.}
        \label{fig:01-traj-Px-Py}
    \end{subfigure}
    \hfill
     \begin{subfigure}{0.49\textwidth}
        \centering
         \includegraphics[width=\textwidth]{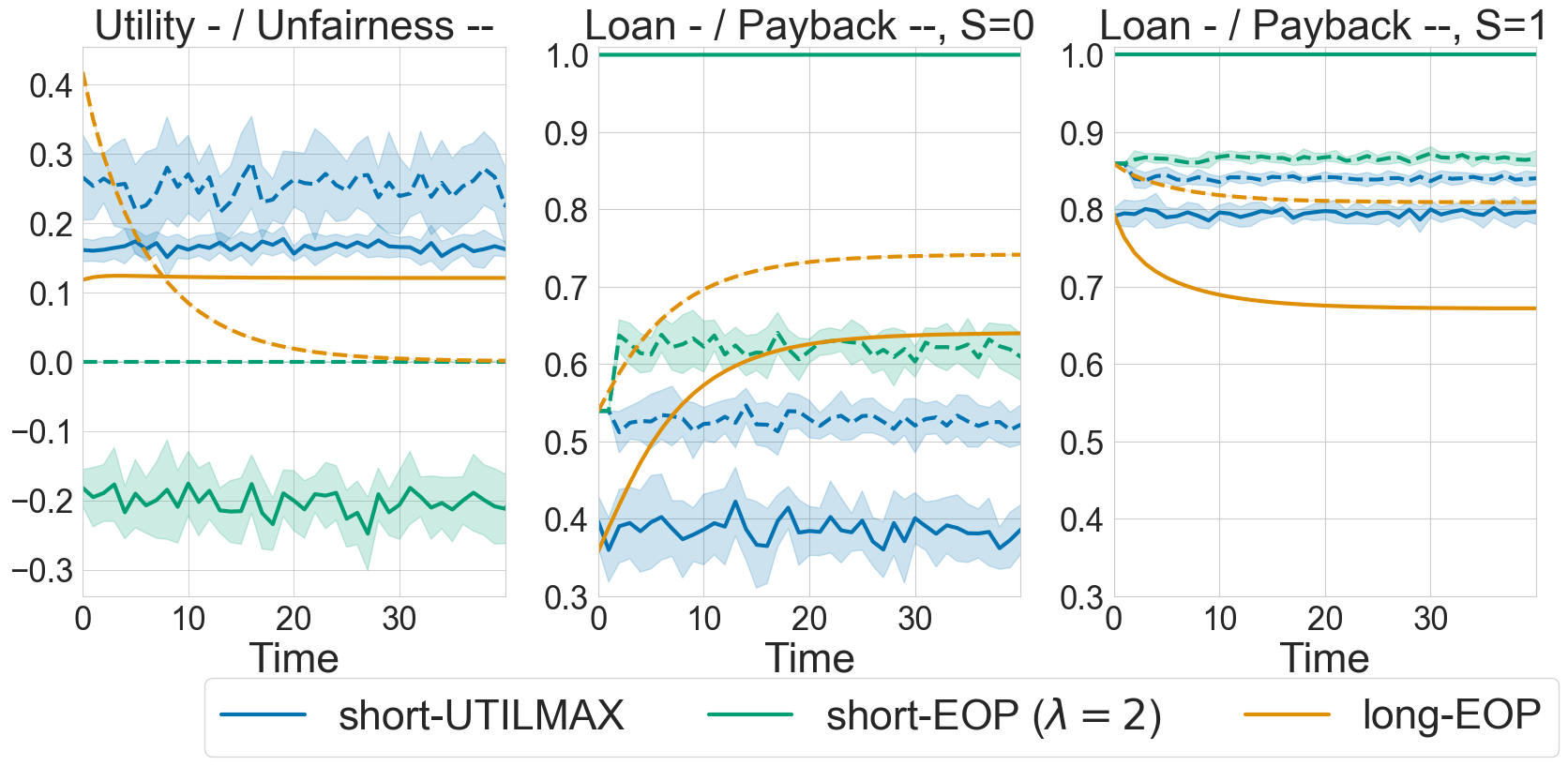}
        \caption{Utility (solid, $\uparrow$), EOP-Unfairness (dashed, $\downarrow$) for short-term-UTILMAX (unfair), short-term-EOP policies (10 seeds), our long-term-EOP policy. Loan (solid) and payback probab. (dashed) per sensitive $S$.}
        \label{fig:06-util-loan}
    \end{subfigure}
    \caption{(a) Convergence independent of initial distribution. (b) Comparison to short-term policies.}
    \vspace{-10pt}
\end{figure}

\subsection{Convergence to Targeted Distribution and Temporal Stability}\label{sec:sim_validation}
\vspace{-5pt}

We demonstrate that a policy derived from an optimization problem based on the general principle converges to a stable steady-state distribution. For setup details see Appendix~\ref{apx:sim}.
%
\paragraph{One-sided Dynamics.}  
One-sided dynamics are characterized by a particular (usually positive) decision leading to changes in a feature distribution, while other decisions do not incur any feature changes. 
Following prior work~\citep{liu2018delayed, damour2020fairness}, we assume in our scenario, that if
an applicant defaults on their loan, their credit score remains the same; if the applicant repays the loan, their credit score is likely to increase. We refer to these dynamics as \onesided.
%
\paragraph{Maximum Utility under EOP-Fairness.} 
%
We now exemplify a long-term target.
Consider a bank that aims to maximize  its
profit ($\utility$) while guaranteeing equal opportunity ($\eopunf$) for loan approval.
Given cost of a positive decision $c$ and a small unfairness level $\epsilon$, we seek for a policy:
\begin{align}\label{op:utilmaxeop}
\begin{split}
     & \maxuitleopstar := \arg_{\policy} \max \utility(\pi;c) \quad \quad \text{subj. to} \quad \eopunf(\pi) \leq \epsilon ; \, \,  \Cconv(\kernelpi),
\end{split}
\end{align}
%
This target has been proposed for fair algorithmic decision-making in static systems~\citep{hardt2016equality}, short-term policies aiming to fulfill this target at each time step have examined in dynamical systems~\citep{zhang2020fair, creager2020causal, damour2020fairness} and has been imposed as long-term target~\cite{wen2021algorithms}.
We redefine this concept as a long-term goal for the stationary distribution to satisfy.

\paragraph{Results.}
%
We run simulations on 10 randomly sampled initial feature distributions $\initialdistribution$, setting $\epsilon=0.01, c=0.8$. Figure~\ref{fig:01-traj-Px-Py} displays the resulting trajectories of the feature distribution for $X_1$ converging to a stationary distribution. For other features see Appendix~\ref{apx:additional-start-distributions}).
We observe that while the initial distribution impacts convergence process and time, the policy consistently converges to a single stationary distribution regardless of starting point.
{The policy found for one population can thus be effectively applied to other populations with different feature distributions, if dynamics and labeling distributions remain unchanged.}
As the outcome of interest $Y$ depends on the features, its distribution converges also to a stationary point.

We now compare our found long-term fair policy to both fair and unfair short-term policies. Figure~\ref{fig:06-util-loan} displays $\utility$ and $\eopunf$. 
Using the initial distribution $\initialdistribution$ from FICO, we solve the optimization problem (\ref{op:utilmaxeop}) for tolerated unfairness $\epsilon = 0.026$. The short-term policies consist of Logistic Regression models for 10 random seeds, which are retrained at each time step; fairness is enforced using a Lagrangian approach ($\lambda=2$). Our policy demonstrates high stability in both utility and fairness compared to short-term policies, which exhibit high variance across time. Note since our policy does not require training, we do not report standard deviation over different seeds. Furthermore, while our policy converges to the same fairness level as the short-term fair policy, it experiences only a marginal reduction in utility compared to the (unfair) utility-maximizing short-term policy. Thus, it does not suffer from a fairness-utility trade-off to the extent observed in the short-term policies.

Figure~\ref{fig:06-util-loan} (middle, right) displays loan $\Pbb(D\!=\!1\!\mid\!S\!=\!s)$ and payback probabilities $\Pbb(Y\!=\!1\!\mid\!S\!=\!s)$ for non-privileged ($S=0$) and privileged ($S=1$) groups.
The short-term fair policy achieves fairness by granting loans to everyone. For the utility-maximizing short-term policy, unfairness arises as 
gap between ability to pay back and loan provision is much smaller for the privileged group, resulting in significantly different loan probabilities between the two groups.
For our long-term policy, we observe that loan provision probabilities converge closely for both groups over time, while the gap between payback probability and loan granting probability remains similar between groups.
Similar to prior research~\citep{wen2021algorithms, yu2022policy}, we observe that our policy achieves long-term objectives, but the convergence phase may pose short-term fairness challenges. In practice, it is essential to assess the potential impact of this on public trust.
%
\subsection{Long-Term Effects of Targeted States}\label{sec:sim_dynamics}
\vspace{-5pt}
%
This section examines the long-term effects of policies and their targeted stationary distributions.
The observations are specific to the assumed dynamics and distributions and
serve as a starting point for a thoughtful reflection on the formulation and evaluation of long-term targets.

\paragraph{Maximum Qualifications.} 
Inspired by~\citep{zhang2020fair}, assume a non-profit organization offering loans. Their goal is to optimize the overall payback ability ($\qualification$) of the population to promote societal well-being. Additionally, they aim to sustain their lending program by prevent non-negative profits ($\utility$) in the long-term.
We thus seek for:
\begin{align}\label{op:maxqual}
\begin{split}
    & \maxqualstar := \arg_{\policy} \max \qualification(\pi) \quad \quad \text{subj. to} \quad \utility(\pi) \geq 0;  \, \, \Cconv(\kernelpi)
\end{split}
\end{align}
%

\paragraph{Two-sided Dynamics.} 
%
In addition to one-sided dynamics, where only positive decisions impact the future, we also consider two-sided dynamics~\citep{zhang2020fair}, where both positive and negative decisions lead to feature changes. We investigate two types of two-sided dynamics.
\begin{wrapfigure}{r}{0.48\textwidth}
    \centering
    \vspace{2pt}
\includegraphics[width=0.48\textwidth]{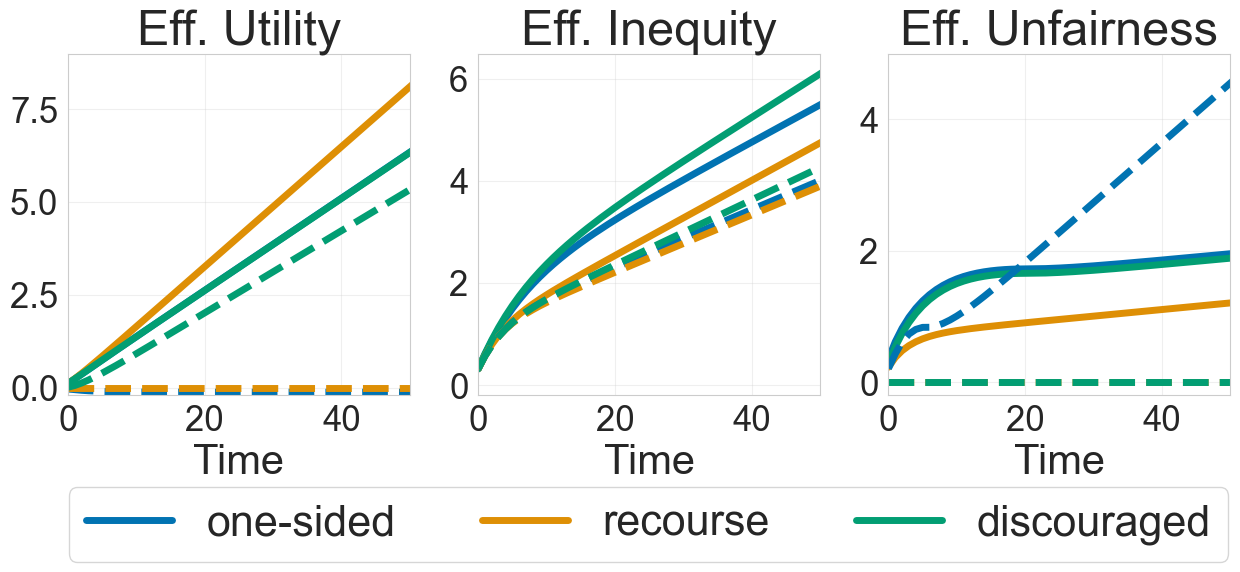}
    \caption{Effective utility $\utility$, inequity $\inequity$ and $\eopunf$ for policies $\maxuitleopstar$ (solid), $\maxqualstar$ (dashed) and different dynamics (colors).}
    \label{fig:04-eff-measures}
    \vspace{-30pt}
\end{wrapfigure}
Under \recourse\ dynamics, individuals receiving unfavorable lending decisions take actions to improve their credit scores, 
facilitated through recourse~\citep{karimi2021causalrecourse} or social learning~\citep{heidari2019effort}.%
In \discouraged\ dynamics, unfavorable lending decisions demotivate individuals, causing a decline in their credit scores. This may happen when individuals cannot access loans for necessary education, limiting their financial opportunities.

\paragraph{Results.}
We solve both introduced optimization for policies
$\maxuitleopstar$(\ref{op:utilmaxeop}) and $\maxqualstar$(\ref{op:maxqual}) with ${c=0.8}$ and ${\epsilon=0.01}$, both subject to convergence constraints $\Cconv$ (irreducibility, aperiodicity), for \onesided, \recourse\ and \discouraged\ dynamics. 
Utilizing the found policies we simulate the feature distribution over 200 time steps, starting from the initial FICO feature distribution. For more details, refer to Appendix~\ref{apx:sim}.
Figure~\ref{fig:04-eff-measures} shows accumulated (effective) measures of utility, inequity and EOP-Unfairness over time.
Across different dynamics, the policies conform with their targets. 
$\maxuitleopstar$ accumulates across dynamics most utility, while $\maxqualstar$ has a small negative cumulative utility due to the imposed zero-utility constraint.
In the \onesided\ scenario, we observe for unfairness different short-term and long-term effects. Up to approx. 40 time steps, $\maxqualstar$ yields lower unfairness than $\maxuitleopstar$, after this point $\maxqualstar$ becomes highly unfair. 
These observations highlight that: dynamics may significantly impact the final outcome of decision policies; when deploying a policy in the long-term small differences in policies can lead to large accumulated effects;
and short term effects may differ from long-term goals.

%% file: sections/08_Discussion.tex
\section{Discussion}\label{sec:discussion}
\vspace{-8pt}
In this section, we discuss key assumptions and limitations. Additional discussion in Appendix~\ref{apx:additional-clarifications}.
\vspace{-4pt}
\paragraph{Limitations of Assumptions.}
The proposed general optimization problem (\ref{op:general}) assumes a time-homogeneous kernel and access to the dynamics defining it.
Although real-world data often change over time, we treat the dynamics as static for a shorter duration, which is plausible, if they rely on bureaucratic~\citep{liu2018delayed} or algorithmic recourse policies~\citep{karimi2020survey}, and if convergence time remains relatively short, as seen in our simulations. 
However, convergence time depends on the dynamics and initial distribution. If the transition probabilities become time-dependent, updating the policy would be necessary.
Transition probabilities for discrete state spaces can be estimated from temporal data~\citep{sherlaw1995estimating, craig2002estimation}, but remains a challenge for continuous state spaces in practice~\citep{duffie2004estimation}. 
Furthermore, few temporal datasets for fair machine learning exist~\citep{mehrabi2019survey}.
Assuming dynamics with expert knowledge is an alternative, but caution is needed as it may lead to
confirmation bias~\citep{nickerson1998confirmation}.

\paragraph{The Case of Non-existence of a Long-Term Fair Policy.}
Consider the case that no solution exists for our problem~(\ref{eq:general-objective}). Then, as argued in \S~\ref{sec:general-goal}, no policy maker with different strategies of finding policies over time would find a solution to the same problem, with the same assumed distributions, dynamics, and constraints.
If a solution to our optimization problem does not exist, this insight may prompt practitioners to explore alternative approaches for long-term fairness, such as non-stationary objectives~\citep{zhang2020fair} or redefining the fair state. Thus, our approach enhances the understanding of system dynamics and long-term fairness.
%

%% file: sections/09_Summary-Outlook.tex
\section{Summary and Outlook}\label{sec:summary-outlook}
\vspace{-5pt}
We have introduced a general problem for achieving long-term fairness in dynamical systems, where algorithmic decisions in one time step impact individuals' features in the next time step, which are consequently used to make decisions. 
 We proposed an optimization problem for identifying a time-independent policy that is guaranteed to converge to a targeted fair stationary state, regardless of the initial data distribution. 
 %
  We model the system dynamics with a time-homogeneous Markov chain and enforce the conditions of the Markov chain convergence theorem to the Markov kernel through policy optimization.
 Our framework can be applied to different dynamics and long-term fair goals as we have shown in a guiding example on credit lending.
 In semi-synthetic simulations, we have shown the effectiveness of policy solutions to converge to targeted stationary population states and illustrated how our approach facilitates the evaluation of different long-term targets. 
Future work lies in applying our framework to a wider range of problems with more complex dynamics, larger (potentially continuous) feature spaces, and multiple sensitive attributes and using more sophisticated optimization methods. 
Future work may also explore the application of our framework to designing social interventions on the transition probabilities~\citep{heidari2019effort,  von2020fairness, mhasawade2021causal}
providing additional insights and solutions for long-term fairness in algorithmic decision-making in dynamical systems.

%% file: sections/acknowledgements.tex
\section*{Acknowledgements}
\vspace{-8pt}
The authors thank Ayan Majumdar and Jonas Klesen and especially Diego Baptista Theuerkauf for providing insightful comments and discussion. 
MR thanks the ELLIS Unit Amsterdam and the German Federal Ministry
of Education and Research (BMBF): Tübingen AI Center, FKZ:
01IS18039B for generous funding support and Max Planck Institute for Intelligent Systems, Tübingen.

%% file: sections/90_Appendix_Markov.tex
\section{Markov Chain Convergence Theorem for General State Spaces}\label{apx:markov}
In this section, we present the Markov convergence theorem for general state spaces, as well as the conditions to satisfy the conditions of the theorem.
We mainly follow the references of \cite{RR04,MT12,AG10,SS21}.

\begin{Not} The following notations will be used.
    \begin{enumerate}
        \item $\Xcal$ denotes a standard measurable space (aka standard Borel space), like $\Xcal=\R^D$ or $\Xcal=\N$, etc. 
        \item We use $\Bcal_\Xcal$ to denote the $\sigma$-algebra of (Borel subsets of) $\Xcal$.
        \item $T:\, \Xcal \dshto \Xcal$ denotes a Markov kernel (aka transition probability) from $\Xcal$ to $\Xcal$, i.e.\ formally a measurable map $T:\, \Xcal \to \Pcal(\Xcal)$ from $\Xcal$ to the space of probability measures over $\Xcal$.
        \item For a point $x \in \Xcal$ and measurable set $A \in \Bcal_\Xcal$ we write $T$ similar to a conditonal probability distribution:
            \begin{align}
            \begin{split}
                 & T(A|x) := T_x(A) \\
                 &:= \text{ probability of $T$ hitting $A$}\\
                 & \text{when starting from point $x$.}
            \end{split}
            \end{align}
        \item We define the Markov kernel $T^0:\, \Xcal \dshto \Xcal$ via: $T^0(A|x) := \I_A(x)$.
        \item We inductively define the Markov kernels $T^n:\, \Xcal \dshto \Xcal$ for $n \in \N_1$ via:
            \begin{align}
            \begin{split}
                & T^n(A|x) := \int_\Xcal T(A|y)\, T^{n-1}(dy|x) 
                \\
                & = \overbrace{(T \circ T \circ \dots \circ T \circ T)}^{n\text{-times}}(A|x).
            \end{split}
            \end{align}
            Note that: $T^1 = T$.
        \item As the sample spaces we consider the product space:
            \begin{align}
                \Omega &:= \prod_{n \in \N_1} \Xcal.
            \end{align}
        \item For $n \in \N_1$ we have the canonical projections: 
            \begin{align}
            \begin{split}
                & X_n:\, \Omega \to \Xcal, \\
                & \qquad \omega=(x_n)_{n \in \N_1} \mapsto x_n =: X_n(\omega).
            \end{split}
            \end{align}
        \item We use $P_x := T^{\otimes \N_1}_x$ to denote the probability measure on $\Omega$ of the homogeneous Markov chain induced by $T$ that starts at $X_0=x$.
            Note that for $n \in \N_1$ the marginal distribution is given by: 
            \begin{align} 
                P_x(X_n \in A) &= T^n(A|x).
            \end{align}
        \item We abbreviate the tuple: $\Xbf :=(X_n)_{n \in \N_1}$. 
            Note that $\Xbf$ is a (homogeneous) Markov chain that starts at $X_0=x$ under the probability distribution $P_x$. We will thus also refer to $\Xbf$ as the (homogeneous) Markov chain corresponding to $T$.
        \item We abbreviate the probability of the Markov chain of ever hitting $A \in \Bcal_\Xcal$ when starting from $x \in \Xcal$ as:
            \begin{align}
                L(A|x) &:= P_x\lp \bigcup_{n \in \N_1} \lC X_n \in A  \rC \rp.
            \end{align}
        \item We abbreviate the probability of the Markov chain hitting $A \in \Bcal_\Xcal$ infinitely often when starting from $x \in \Xcal$ as:
            \begin{align}
                 Q(A|x):= P_x\lp \lC X_n \in A \text{ for infinitely many } n \in \N_1  \rC \rp.
            \end{align}
        \item We abbreviate the expected number of times the Markov chain hits $A \in \Bcal_\Xcal$ when starting from $x \in \Xcal$ as:
            \begin{align}
            \begin{split}
                & U(A|x) := \sum_{n \in \N_1} T^n(A|x) = \E_x[\eta_A], \\
                & \eta_A:=\sum_{n \in \N_1} \I_A(X_n).
            \end{split} 
            \end{align}
    \end{enumerate}
\end{Not}

\begin{Def}[Irreducibility]
$T$ is called \emph{irreducible} if there exists a non-trivial $\sigma$-finite measure $\phi$ on $\Xcal$ such that for $A \in \Bcal_\Xcal$ we have the implication:
            \begin{align}
                \phi(A) >0 \quad \implies \quad \forall x \in \Xcal. \quad L(A|x) > 0. \label{eq:irr}
            \end{align}
\end{Def}

The statement from \cite{MT12} Prp.\ 4.2.2 allows for the following remark.

\begin{Rem}[Maximal irreducibility measure]
    If $T$ is irreducible then there always exists a non-trivial $\sigma$-finite measure $\psi$ that is maximal (in the terms of absolute continuity) among all those $\phi$ with property \ref{eq:irr}.
    Such a $\psi$ is unique up to equivalence (in terms of absolute continuity) and is called a \emph{maximal irreducibility measure of $T$}.
    For such a $\psi$ we introduce the notation:
    \begin{align}
        \Bcal_\Xcal^T &:=\lC A \in \Bcal_\Xcal \st \psi(A) >0 \rC.
    \end{align}
    Note that $\Bcal_\Xcal^T$ does not depend on the choice of a maximal irreducibility measure $\psi$ due to their equivalence. With this notation we then have for irreducible $T$:
            \begin{align}
                A \in \Bcal_\Xcal^T \quad \implies \quad \forall x \in \Xcal. \quad L(A|x) >0.
            \end{align}   
\end{Rem}

\begin{Def}[Harris recurrence]
$T$ is called \emph{Harris recurrent} if $T$ is irreducible and we have the implication:
            \begin{align}
                A \in \Bcal_\Xcal^T \quad \implies \quad \forall x \in \Xcal. \quad L(A|x) =1.
            \end{align}        
\end{Def}

\begin{Def}[Invariant probability measures]
    An \emph{invariant probability measure (ipm)} of $T$ is a probability measure $\mu$ on $\Xcal$ such that:
    \begin{align}
        T \circ \mu & = \mu.
    \end{align}
    On measurable sets this can equivalently be re-written as:
    \begin{align}
      \forall A \in \Bcal_\Xcal.\qquad  \int_\Xcal T(A|x) \, \mu(dx) & = \mu(A).
    \end{align}
\end{Def}

\begin{Rem}
    Note that a general Markov kernel $T$ can have either no, exactly one or many invariant probability measures.
\end{Rem}


For irreducible $T$ we have the following results from \cite{MT12} Prp.\ 10.1.1, Thm.\ 10.4.4, 18.2.2,
concerning existence and uniqueness of invariant probability measures.

\begin{Thm}[Existence and uniqueness of invariant probability measures] 
    Let $T$ be irreducible.     \begin{enumerate}
        \item Then $T$ has at most one invariant probability measure $\mu$; and:
        \item the following are equivalent:
    \begin{enumerate}
        \item $T$ has an invariant probability measure $\mu$;
        \item the following implication holds for $A \in \Bcal_\Xcal$:
            \begin{align}
            \begin{split}
                 & A \in \Bcal_\Xcal^T \quad \implies \quad \forall x \in \Xcal. \\
                 & \quad   \limsup_{n \to \infty} T^n(A|x) >0.
            \end{split}
            \end{align}
    \end{enumerate}
    \end{enumerate}
\end{Thm}

We have the following properties of invariant probability measures for irreducible $T$. 
These are cited from \cite{MT12} Thm.\ 9.1.5, Prp.\ 10.1.1, Thm.\ 10.4.4, 10.4.9, 10.4.10, and, \cite{SS21} Prp.\ A.1, Lem.\ 3.2.

\begin{Thm}[Properties of irreducible Markov kernels with invariant probability measures]
        Let $T$ be irreducible with invariant probability measure $\mu$.    Then the following statements hold:
        \begin{enumerate}
        \item $\mu$ is a maximal irreducibility measure for $T$.
        \item $\mu$ satisfies the following condition for every $A \in \Bcal_\Xcal^T$ and $B \in \Bcal_\Xcal$:
    \begin{align}
        \mu(B) = \int_A \E_x\lB \sum_{n=1}^{\tau_A} \I[X_n \in B] \rB\, \mu(dx), \quad \tau_A := \inf \lC n \in \N_1 \st X_n \in A \rC.
    \end{align}
\item  There exists a measurable set $\Hcal \in \Bcal_\Xcal^T$ with $\mu(\Hcal)=1$ such that: 
     \begin{align}
         \forall x \in \Hcal. \quad T(\Hcal|x) =1,
     \end{align}
     $T$ restricted to $\Hcal$, $T:\, \Hcal \dshto \Hcal$, is well-defined and Harris recurrent (with invariant probability measure $\mu$). 
    \end{enumerate}
\end{Thm}

\begin{Def}[Aperiodicity]
    Let $T$ be irreducible. Then $T$ is called:
    \begin{enumerate}
        \item \emph{periodic} if there exists $d \ge 2$ pairwise disjoint sets 
            $A_1,\dots,A_d \in \Bcal_\Xcal^T$, such that for every $j =1, \dots, d$, we have:
            \begin{align}
                \forall x \in A_j.\quad  T(A_{j+1(\mathrm{mod}\,d)}|x) =1;
            \end{align}
        \item \emph{aperiodic} if $T$ is not periodic.
    \end{enumerate}
\end{Def}

With these notation we have the following convergence theorems, see \cite{MT12} Thm.\ 13.3.3,  17.0.1,  and, \cite{SS21} Thm.\ 2.16, 2.17, Assm.\ 2.12, Prp.\ 2.2.

\begin{Thm}[Strong Markov chain convergence theorem]
    \label{thm:strong-mcct}
    Let $\mu$ be a probability measure on $\Xcal$.
    Then the following are equivalent:
    \begin{enumerate}
        \item $T$ is aperiodic and Harris recurrent and $\mu$ is an invariant probability measure for $T$.
        \item For every $x \in \Xcal$ we have the convergence in total variation norm:
    \begin{align}
        \lim_{n \to \infty} \TV(T^n_x,\mu) &= 0.
    \end{align}
    \end{enumerate}
    Furthermore, if this is the case, then for every
    $g \in L^1(\mu)$ and every starting point $x \in \Xcal$ we have the convergences:
    \begin{align}
        \lim_{n \to \infty} \frac{1}{n} \sum_{k=1}^n g(X_k) &= \E_\mu[g]  \quad P_x\text{-a.s.}
    \end{align}
\end{Thm}

\begin{Thm}[Markov chain convergence theorem]
    Let $\mu$ be a probability measure on $\Xcal$.
    Then the following are equivalent:
    \begin{enumerate}
        \item $T$ is aperiodic and irreducible and $\mu$ is an invariant probability measure for $T$.
        \item For every $x \in \Xcal$ we have:
    \begin{align}
        \lim_{n \to \infty} \TV(T^n_x,\mu) < 1,
    \end{align}
    and, for $\mu$-almost-all $x \in \Xcal$ we have the convergence in total variation norm:
    \begin{align}
        \lim_{n \to \infty} \TV(T^n_x,\mu) &= 0.
    \end{align}
    \end{enumerate}
    Furthermore, if this is the case, then for every
    $g \in L^1(\mu)$ and $\mu$-almost-all starting points $x \in \Xcal$ we have the convergences:
    \begin{align}
        \lim_{n \to \infty} \frac{1}{n} \sum_{k=1}^n g(X_k) &= \E_\mu[g]  \quad P_x\text{-a.s.}
    \end{align}
\end{Thm}

We now want to investigate under which conditions we can achieve irreduciblity, aperiodicity or Harris recurrence. We first cite the results of \cite{AG10} Thm.\ 1 and Cor.\ 1.

\begin{Thm}[Harris recurrence via irreducibility and density]
    Let $T$ be irreducible with invariant probability measure $\mu$. 
    Further, assume that $T$ has a density w.r.t.\ an irreducibility measure $\phi$, i.e.:
    \begin{align}
        T(A|x) & = \int_A t(y|x) \, \phi(dy),
    \end{align}
    with a jointly measurable $t:\,\Xcal \times \Xcal \to \R_{\ge 0}$.
    Then $\phi$ is a maximal irreducibility measure for $T$, $\mu$ has a strictly positive density w.r.t.\ $\phi$ and $T$ is Harrris recurrent.
\end{Thm}

\begin{Cor}[Harris recurrence via irreducibility and Metropolis-Hastings form]
    Let $T$ be irreducible with invariant probability measure $\mu$.
    Further, assume that $T$ is of Metropolis-Hastings form w.r.t.\ an irreducibility measure $\phi$:
    \begin{align}
         T(A|x) = (1-a(x))\cdot \I_A(x) + \int_A a(y|x) \cdot q(y|x) \,\phi(dy),
    \end{align}
    with jointly measurable $a, q:\,\Xcal \times \Xcal \to \R_{\ge 0}$ and $a(x) > 0$ for 
    every $x \in \Xcal$. Note that: $a(x) = \int a(y|x) \cdot q(y|x) \,\phi(dy)$.
    Then $\phi$ is a maximal irreducibility measure for $T$, $\mu$ has a strictly positive density w.r.t.\ $\phi$ and $T$ is Harrris recurrent.
\end{Cor}

We now have all ingredients to derive the following criteria for the strong Markov chain convergence theorem \ref{thm:strong-mcct} to apply:

\begin{Cor}[Criterion for convergence via positive density]
Let $\phi$ be a non-trivial $\sigma$-finite measure on $\Xcal$ such that $T$ has a strictly positive jointly measurable density $t:\, \Xcal \times \Xcal \to \R_{> 0}$  w.r.t.\ $\phi$:
    \begin{align}
        T(A|x) &= \int_A t(y|x)\, \phi(dy),
    \end{align}
then $T$ is irreducible, aperiodic and $\phi$ is a maximal irreducibility measure for $T$.

If, furthermore, $T$ has an invariant probability measure $\mu$ then $\mu$ has a strictly positive density w.r.t.\ $\phi$, 
$T$ is Harris recurrent and the strong Markov chain convergence theorem \ref{thm:strong-mcct} applies.
\end{Cor}

\begin{Cor}[Criterion for convergence via positive Metropolis-Hastings form]
    Let $\mu$ be an invariant probability measure of $T$.
    Further, assume that $T$ is of Metropolis-Hastings form w.r.t.\ a non-trivial $\sigma$-finite measure $\phi$:
    \begin{align}
        T(A|x) = (1-a(x))\cdot \I_A(x) + \int_A a(y|x) \cdot q(y|x) \,\phi(dy),
    \end{align}
    with strictly positive jointly measurable $a, q:\,\Xcal \times \Xcal \to \R_{> 0}$ such that
    for every $x \in \Xcal$ we have that: 
    \begin{align}
        a(x) &:= \int a(y|x) \cdot q(y|x) \,\phi(dy) \stackrel{!}{\in} (0,1).
    \end{align}
    Then $\phi$ is a maximal irreducibility measure for $T$, $\mu$ has a strictly positive density w.r.t.\ $\phi$, $T$ is aperiodic, Harrris recurrent and the strong Markov chain convergence theorem \ref{thm:strong-mcct} applies.
\end{Cor}

\begin{Cor}[Criterion for convergence on countable spaces]
    Let $\Xcal$ be a countable space, i.e.\ finite or countably infinite.
    Let $T$ be irreducible with invariant probability measure $\mu$ such that for 
    all $x \in \Xcal$ with $\mu(\lC x \rC) >0$ we also have $T(\lC x\rC|x) >0$.
    Then $T$ is aperiodic and Harris recurrent and the strong Markov chain convergence theorem \ref{thm:strong-mcct} applies.
\end{Cor}

%% file: sections/96_Appendix_clarifications.tex
\section{{Additional Clarifications and Discussion}}\label{apx:additional-clarifications}

In this section, we provide additional clarifications and discussion. 

\subsection{Definition of Long-term Fairness}\label{apx:clarification-definition-long-term-fairness}
We provide an overview of how prior research’s fairness formulations relate to our definitions of long-term fair targets.

First, our framework aims to attain a state of long-term fairness. This entails that fairness formulations should be met in the long term and, importantly, once achieved, maintained consistently.
Our goal differs fundamentally from approaches that aim to fulfill fairness at each time step. In this regard, \citep{damour2020fairness} compare agents optimizing for short-term goals - e.g., a profit-maximization agent to an equality of opportunity fair agent and  measure the long-term (in)equality of  the initial credit score distribution across groups - without imposing it on the agents.

Prior work on long-term fairness  introduces parity of return~\citep{chi2022towards}, which requires equal discounted rewards accumulated by the decision-maker over time, where the reward could be defined as the ratio between true positive and overall positive decisions. 
\cite{wen2021algorithms} define long-term demographic parity (equal opportunity) as asking the cumulative expected individual rewards to be on average  equal for (qualified members of) demographic groups. 
\citep{yin2023long} aim to maximize the accumulated reward subject to accumulated unfairness (utility) constraint in a finite time horizon. The reward combines true positive and true negative rates, while the authors consider different (un)fairness measures: demographic parity, equal opportunity, and equal qualification rate.
\citep{yu2022policy} formulate a (short-term) fairness metric (e.g., equality of opportunity) as a function of the state and increase its enforcement over time.

Our framework provides the capability to enforce these fairness and reward considerations, specifically we allow for feature complex objective functions (see \S~\ref{sec:goals-profit}) as well as imposing feature (qualification) equality \S~\ref{sec:goals-distributional-fairness}) and group fairness criteria in the long-term (see \S~\ref{sec:goals-predictive-fairness}) for infinite time-horizons.
Note that the formulation of a fair state is not limited to the possible fairness objectives and constraints discussed in \S~\ref{sec:goals}. Rather, we exemplify in that section that our framework can capture fairness objectives well-established in prior work (in addition to the above cited:~\cite{zhang2020fair, liu2018delayed, dwork2012fairness, hardt2016equality}).

 \subsection{Assumption of Known or Estimatable Dynamics}
Our work takes a structured approach by separating the estimation problem (of the Markov kernel i.e., the dynamics) from the policy learning process. We recognize that the estimation problem itself is a significant challenge and requires careful attention and, as commented in \S~\ref{sec:discussion}, is the subject of a different line of active research and thus outside the scope of this paper.  

The quality of the dynamics estimation heavily relies on the quality and quantity of the available temporal data, the complexity of the environment, and the estimation methods (as it does, e.g., for model-based reinforcement learning). Estimation of dynamics / Markov kernels is an active research field~\cite{sherlaw1995estimating, craig2002estimation,wu2018deep, sun2019learning} and our method can benefit from the advances made in the field. If temporal data is available, estimating dynamics may even prove to be faster and more data-efficient than learning them through interactions. We exemplify estimating dynamics in additional results in Appendix~\ref{apx:additional-estimations}.

Further, within our framework and application, dynamics we describe consequences of decisions on individuals’ features. The dynamics in the lending example of our experiments are determined by the credit score maker’s policy on how scores are updated in response to (un)paid credits. Though our framework is not limited to this, dynamics - themselves depending on a statistical/rule-based/ML model - may be accessible or much simpler to estimate than complex human behavior.

\subsection{Existence of a Fair Stationary Distribution}
Our approach also serves to determine whether a stationary distribution exists. In situations where a fair policy does indeed exist, our optimization problem (OP) is designed to effectively discover it. If a solution to our optimization problem does not exist, it implies that alternative methods (including, e.g., reinforcement learning), would also not find a policy inducing and maintaining the targeted fair stationary distribution under the same modeling assumptions. This stems from the fact that if the current state is fair, any alternative approach would still need to address the stationary equation (\ref{eq:general-objective}) to maintain that state. 
This discovery can offer valuable insights to practitioners, prompting them to explore different perspectives on long-term fairness. For instance, this might involve revising non-stationary long-term fairness objectives, such as addressing oscillating long-term behaviors \citep{zhang2020fair}. Alternatively, practitioners could consider redefining the targeted fair state that allows for stationary. By shedding light on these possibilities, our approach contributes to a deeper understanding of the dynamics and long-term fairness considerations.

\subsection{Choice of Dataset} 
Our current experiment focuses on a single simulation setup, specifically centered around loan repayment. At the same time, we provide results for varying dynamics and initial distributions, essentially simulating different datasets of the same generative model. Note also that we provide an example of how the framework can be applied to a different generative model in Appendix~\ref{apx:example}. 
Finally, focusing on a single generative model~\citep{zhang2020fair} and a single guiding example is in line with prior published work~\citep{zhang2020fair, liu2018delayed, creager2020causal, wen2021algorithms} with the loan example used widely by previous work on long-term fairness~\citep{damour2020fairness, liu2018delayed, creager2020causal, wen2021algorithms, yu2022policy}.

\subsection{Opportunities and Limitations of Time-invariant Policies}
Our framework yields a single fixed, i.e., time-invariant policy. 
When the dynamics are constant, and policy learning and estimation of the dynamics occur simultaneously (as in reinforcement learning), then the learned policy requires frequent updates as more data becomes available. Our paper takes a different approach by separating the estimation problem (of the Markov kernel i.e., the dynamics) from the policy learning process and therefore does not require updating the policy. We believe that this holds several advantages, particularly in terms of predictability and trustworthiness. A fixed policy provides a consistent decision-making framework that stakeholders can anticipate and understand contributing to trustworthiness. In addition, a fixed policy simplifies operational processes, such as implementation and maintenance efforts, potentially leading to more efficient and effective outcomes.

When the dynamics vary with time, we can no longer rely on a single time-invariant policy for an infinite time horizon. If, however, the changes are slow and the dynamics remain constant within certain time intervals, our approach remains effective within the time intervals. Whenever the dynamics change, our approach would require re-estimating the dynamics and solving the optimization problem again to obtain a new policy. In this way, our method adapts to changing conditions and maintains its effectiveness over time. However, when dynamics change rapidly, the adaptability of any method is limited.

\subsection{Modeling Choice}
Our intention in developing a framework for long-term fair policy learning is to provide a versatile approach that could be applied across various contexts. While models serve as simplified representations of complex systems, they allow us to analyze phenomena otherwise incomprehensible. Our choice of utilizing Markov Chains as a modeling tool is a reflection of this principle. Markov Chains are chosen for their wide application in understanding dynamic processes. For example, the field of Reinforcement Learning relies on Markov Decision Processes (MDPs), a specific kind of Markov Chain. The proposed modeling framework can indeed be adapted to a variety of different scenarios and we provide an example of a different scenario / generative model in {Appendix~\ref{apx:example}.}

%% file: sections/91_Appendix_long-term-targets.tex
\section{On Long-term Targets}\label{apx:goals}
In this section, we provide additional details regarding the targeted fair states introduced in \S~\ref{sec:goals}. 

\subsection{On Minimax Objectives}
In \S~\ref{sec:goals-distributional-fairness}, it was mentioned that egalitarian distributions may not always be efficient, and there are cases where minimizing the maximum societal risk is more desirable to prevent unnecessary harm. We elaborate on this concept in the following.
While egalitarian allocations can align with societal values, they are generally considered Pareto inefficient~\citep{pazner1975pitfalls}. In certain scenarios, policy-makers may be interested in minimizing the maximum risk within a society~\citep{barsotti2022minmax}. This approach aims to prevent unnecessary harm by reducing the risk for one group without increasing the risk for another~\citep{martinez2020minimax}.
For instance, in the context of hiring, instead of equalizing the group-dependent repayment rates $\qualification(\policy,s)$, a policy-maker may be interested in minimizing the maximum default risk $1-\qualification(\policy,s)$ across groups. In other words, their objective could be $\objective:=\min_s - (1-\qualification(\policy,s))$, rather than aiming for equal default or repayment rates.

\subsection{Policy constraints}
In \S~\ref{sec:goals-predictive-fairness}, we mentioned that it is possible to incorporate constraints on the type of policy being searched for. 
These constraints could be put on the policy independent of the stationary distribution. 
We provide an example here.
If the features exhibit a monotonic relationship, where higher values of $X_t$ tend to result in a higher probability of a positive outcome of interest $\labelfunctone$, we may also be interested in a monotonous policy. A monotonous policy assigns higher decision probabilities as $X_t$ increases.
 In such cases, we can impose the additional constraint ${\policy(k, s) \geq \policy(x,s) , \forall k \geq x, s }$.

%% file: sections/92_Appendix_Simulation.tex
\section{Simulation Details}\label{apx:sim}
In this section, we present the details of the experiments and simulations in \S~\ref{sec:simulations}.

\subsection{Solving the Optimization Problem}
Our framework can be thought of as a three-step process. First, just as previous work on algorithmic fairness empowers users to choose fairness criteria, our framework allows users to define the characteristics of a fair distribution applicable in their decision-making context (see \S~\ref{sec:goals}). The second step involves transforming the definition of fair characteristics into an optimization problem (OP). The third step consists of solving the OP. Given the nature of our optimization problem, which is linear and constraint-based, we can employ any efficient black-box optimization methods for this class of problems.
Note that the OP seeks to find a policy $\pi$ that induces a stationary distribution $\mu$, which adheres to the previously defined fairness targets. As detailed in \S~\ref{sec:simulations}, in the search of $\pi$, we first compute group-dependent kernel $T_{\pi}^s$, which is a linear combination of assumed/estimated dynamics and distributions and $\pi$. We then compute the group-dependent stationary distribution $\mu_{\pi}^s$ via eigendecomposition.

\paragraph{Solving the Optimization Problem for Finite State Spaces} 
In our guiding example and the corresponding simulation, we consider a time-homogeneous Markov chain $(\statespace, \kerneldef)$ with a finite state space $\statespace$ (e.g., credit score categories). Consequently, the convergence constraints $\Cconv$ are determined by the \emph{irreducibility} and \emph{aperiodicity} properties of the corresponding Markov kernel (see \S~\ref{sec:background}). 

Recall from Def.~\ref{def:irreducibility} that a time-homogeneous Markov chain is considered \emph{irreducible} if, for any two states $\statefrom, \stateto \in \statespace$, there exists a 
$t> 0$ such that $\kerneldef^t(\statefrom, \stateto) > 0$, where ${\kerneldef^t(\statefrom, \stateto)} = {\Pbb(\State_{t}=\stateto \mid \State_0=\statefrom)}$ represents the probability of going from $\statefrom$ to $\stateto$ in $t$ steps. 

To ensure irreducibility in our optimization problem, we impose the condition $\sum_{i = 1}^{n} \kerneldef^n > \mathbf{0}$, where $n = |\statespace|$ is the number of states and $\mathbf{0}$ denotes the matrix with all entries equal to zero. We can demonstrate that this implies irreducibility through a proof by contradiction: Suppose that $\sum_{i = 1}^{n} \kerneldef^n > \mathbf{0}$, but for all $t \in \{1, 2, \ldots, n\}$, we have $\kerneldef^t(\statefrom, \stateto) = 0$ for all $\statefrom$ and $\stateto$. Then $\sum_{t=1}^n \kerneldef^n = 0$, which contradicts the initial condition. Consequently, if $\sum_{i = 1}^{n} \kerneldef^n > \mathbf{0}$, it follows that there exists a 
$t> 0$ such that $\kerneldef^t(\statefrom, \stateto) > 0$.

To satisfy \emph{aperiodicity} in our optimization we require that the diagonal elements of the transition matrix are greater than zero: $\kerneldef(\statefrom, \statefrom) > 0$ for all $\statefrom$, where $\kerneldef(\statefrom, \statefrom)$ represents the diagonal elements of the Markov kernel $\kerneldef$.
Recall from Def.~\ref{def:aperiodicity}, that we denote ${R(z)=\left\{t \geq 1: \kerneldef^t(\statefrom, \statefrom)>0\right\}}$ to be the set of return times from $\statefrom \in \statespace$, where $\kerneldef^t(z,z)$ represents the probability of returning to state $x$ after $t$ steps. 
The Markov chain is aperiodic if and only if the greatest common divisor (gcd) of $R(\statefrom)$ is equal to 1: $gcd(R(z))=1$ for all $\statefrom$ in $\statespace.$
If $\kerneldef^1(\statefrom,\statefrom) > 0$ for all $\statefrom$, then $t=1$ is in $R(\statefrom)$, which means that the gcd of $R(\statefrom)$ is equal to 1.

Following Theorem~\ref{thm:markov-convergence}, a sufficient condition for convergence to the unique stationary distribution is the positivity of the transition matrix $\kerneldef$, where all elements are greater than zero. Therefore, if we assume the transition matrix to be positive, we do not need to impose the \emph{irreducibility} and \emph{aperiodicity} constraints mentioned above. In our experiments, for the sake of simplicity, we ensure that the transition matrix $\kerneldef$ is positive, meaning that all its elements are greater than zero. Specifically, in our guiding example, this assumption implies that we assume $\dynamicfunctall > 0$ for all $d, s, y, x, k$, while FICO data already yields $\labelfunctall > 0$ for all $y, x, s$.

We compute the \emph{stationary distribution} $\stationary$ using eigendecomposition. Recall from Definition~\ref{def:stationary} that a stationary distribution of a time-homogeneous Markov chain $(\statespace, \kerneldef)$ is a probability distribution $\stationary$ such that $\stationary = \stationary \kerneldef$. More explicitly, for every $\stateto \in \statespace$, the following needs to hold: ${\stationary(\stateto) = \sum_{\statefrom} \stationary(\statefrom) \cdot \kerneldef(\statefrom,\stateto)}$.
If the transition matrix $\kerneldef$ is positive, $\stationary = \stationary \kerneldef$ implies that $\stationary$ is the eigenvector of $\kerneldef$ corresponding to eigenvalue 1. 
We then solve for the stationary distribution $\stationary$ using linear algebra.

\paragraph{SLSQP Algorithm}
We solve optimization problems (\ref{op:utilmaxeop}) and (\ref{op:maxqual}) using the Sequential Least Squares Programming (SLSQP) method~\cite{kraft1988software}. 
SLSQP is a method used to minimize a scalar function of multiple variables while accommodating bounds, equality and inequality constraints and can be used for solving both linear and non-linear constraints.
The algorithm iteratively refines the solution by approximating the objective function and constraints using quadratic model.

Specifically, SLSQP is designed to minimize scalar functions of one or more variables. In our case we are maximizing utility ($\maxuitleopstar$) or qualifications ($\maxqualstar$) and searching for $\Pbb(D=1 \mid X=x, S=s)$ for all $x$ and $s$, which are with $|X|=4$ and $|S|=2$, a total of $8$ variables. Further, SLSQP can handle optimization problems with variable bounds. In our case, we set a minimum bound of 0 and a maximum bound of 1 as we are seeking for probabilities $\Pbb(D=1 \mid X=x, S=s)$ for all $x$ and $s$. SLSQP can also handle both linear and non-linear equality and inequality constraints. In our example, where the state space is finite (i.e., $X$ is categorical), all constraints are linear inequality or equality constraints. Finally, SLSQP uses a sequential approach, which means it iteratively improves the solution by solving a sequence of subproblems. This approach often converges efficiently, even for non-convex and non-linear optimization problems.

We use the SLSQP solver from scikit-learn\footnote{\tiny{\url{https://docs.scipy.org/doc/scipy/reference/optimize.minimize-slsqp.html}}}~\citep{pedregosa2011scikit} with step size $\text{eps} \approx 1.49 \times 10^{-10}$ and a max. number of iterations $200$ and initialize the solver (warm start) with a uniform policy where all decisions are random, i.e., $\policyone=0.5 \, \forall x,s$.

\subsection{Assumed Dynamics}\label{apx:dynamics}
We now provide details about the assumed dynamics.  
Note that in our guiding example, we assume binary $s, y, d\in \{0,1\}$ and four credit categories, i.e., we have $n = |\Xcal| = 4$ states. For simplicity we assume the following notation: $T_{s d y}:= \dynamicfunctall$. $T_{sdy}$ is a $n\times n$ transition matrix that describes the Markov chain, where the rows and columns are indexed by the states, and $T_{sdy}(x, k)$, i.e., the number in the $x$-th row and $k$-th column, gives the probability of going to state $X_{t+1} =  k$ at time $t + 1$, given that it is at state $X_{t} = x$ at time $t$ and given that $S=s$, $D_t=d$, $Y_t=y$.

\paragraph{One-sided Dynamics.}
For all one-sided dynamics (in \S~\ref{sec:simulations} and \ref{apx:additional-speed}) we assume: 

\begin{align}\label{dyn:one-sided}
\begin{split}
   & T_{000} = T_{001} = T_{100} = T_{101}\\
   &= \begin{bmatrix}
0.9 & 0.03333 & 0.03333 & 0.03333 \\
0.03333 & 0.9 & 0.03333 & 0.03333 \\
0.03333 & 0.03333 & 0.9 & 0.03333 \\
0.03333 & 0.03333 & 0.03333 & 0.9 \\
\end{bmatrix}\\
& T_{110} = T_{010} \\
&= \begin{bmatrix}
0.9 & 0.9 & 0.9 & 0.9 \\
0.03333 & 0.03333 & 0.03333 & 0.03333 \\
0.03333 & 0.03333 & 0.03333 & 0.03333 \\
0.03333 & 0.03333 & 0.03333 & 0.03333 \\
\end{bmatrix} 
\end{split}
\end{align}

\paragraph{One-sided General.} For the one-sided dynamics in \S~\ref{sec:sim_validation} we additionally assume dynamics $T_{s d y}$ that depend on the sensitive attribute in addition to (\ref{dyn:one-sided}):
\begin{align*}
T_{111} = \begin{bmatrix}
0.53333 & 0.03333 & 0.03333 & 0.03333 \\
0.4 & 0.53333 & 0.03333 & 0.03333 \\
0.03333 & 0.4 & 0.53333 & 0.03333 \\
0.03333 & 0.03333 & 0.4 & 0.9 \\
\end{bmatrix}\\
T_{011} = \begin{bmatrix}
0.33333 & 0.03333 & 0.03333 & 0.03333 \\
0.6 & 0.33333 & 0.03333 & 0.03333 \\
0.03333 & 0.6 & 0.33333 & 0.03333 \\
0.03333 & 0.03333 & 0.6 & 0.9 \\
\end{bmatrix}
\end{align*}

\paragraph{One-sided Slow.} For the one-sided \slow\ dynamics with results presented in \ref{apx:additional-speed}, we assume the following group-independent dynamics $T_{s d y}$ in addition to (\ref{dyn:one-sided}):
\begin{align*}
T_{011} = T_{111} = \begin{bmatrix}
0.53333 & 0.03333 & 0.03333 & 0.03333 \\
0.4 & 0.53333 & 0.03333 & 0.03333 \\
0.03333 & 0.4 & 0.53333 & 0.03333 \\
0.03333 & 0.03333 & 0.4 & 0.9 \\
\end{bmatrix}
\end{align*}

\paragraph{One-sided Medium.}  For the one-sided \medium\ dynamics with results presented in \ref{apx:additional-speed}, we assume the following group-independent dynamics $T_{s d y}$ in addition to (\ref{dyn:one-sided}):
\begin{align*}
T_{011} = T_{111} = \begin{bmatrix}
0.33333 & 0.03333 & 0.03333 & 0.03333 \\
0.6 & 0.33333 & 0.03333 & 0.03333 \\
0.03333 & 0.6 & 0.33333 & 0.03333 \\
0.03333 & 0.03333 & 0.6 & 0.9 \\
\end{bmatrix}
\end{align*}

\paragraph{One-sided Fast.}  For the one-sided \fast\ dynamics with results presented in \ref{apx:additional-speed}, we assume the following group-independent dynamics $T_{s d y}$ in addition to (\ref{dyn:one-sided}):
\begin{align*}
\begin{split}
& T_{011} = T_{111} \\
& = \begin{bmatrix}
0.13333 & 0.03333 & 0.03333 & 0.03333 \\
0.8 & 0.13333 & 0.03333 & 0.03333 \\
0.03333 & 0.8 & 0.13333 & 0.03333 \\
0.033335 & 0.03333 & 0.8 & 0.9 \\
\end{bmatrix}  
\end{split}
\end{align*}

\paragraph{Two-sided Recourse Dynamics.}
For \recourse\ dynamics we assume the following dynamics $T_{s d y}$. Specifically, we assume that dynamics are the same for both sensitive groups.

\begin{align*}
T_{000} &= T_{001} = \begin{bmatrix}
0.7 & 0.03333 & 0.03333 & 0.03333 \\
0.23333 & 0.7 & 0.03333 & 0.03333 \\
0.03333 & 0.23333 & 0.7 & 0.03333 \\
0.03333 & 0.03333 & 0.23333 & 0.9 \\
\end{bmatrix} \\
T_{100} &= T_{101} = \begin{bmatrix}
0.5 & 0.03333 & 0.03333 & 0.03333 \\
0.43333 & 0.5 & 0.03333 & 0.03333 \\
0.03333 & 0.43333 & 0.5 & 0.03333 \\
0.03333 & 0.03333 & 0.43333 & 0.9 \\
\end{bmatrix} \\
T_{010} &= T_{011} = \begin{bmatrix}
0.9 & 0.9 & 0.9 & 0.9 \\
0.03333 & 0.03333 & 0.03333 & 0.03333 \\
0.03333 & 0.03333 & 0.03333 & 0.03333 \\
0.03333 & 0.03333 & 0.03333 & 0.03333 \\
\end{bmatrix} \\
T_{110} &= T_{111} = \begin{bmatrix}
0.33333 & 0.03333 & 0.03333 & 0.03333 \\
0.6 & 0.33333 & 0.03333 & 0.03333 \\
0.03333 & 0.6 & 0.33333 & 0.03333 \\
0.03333 & 0.03333 & 0.6 & 0.9 \\
\end{bmatrix}
\end{align*}

\paragraph{Two-sided Discouraged Dynamics.}
For \discouraged\ dynamics we assume the following dynamics $T_{s d y}$. Specifically, we assume that dynamics are the same for both sensitive groups.

\begin{align*}
T_{000} &= T_{001} = \begin{bmatrix}
0.9 & 0.63333 & 0.13333 & 0.03333 \\
0.03333 & 0.3 & 0.53333 & 0.23333 \\
0.03333 & 0.03333 & 0.3 & 0.43333 \\
0.03333 & 0.03333 & 0.03333 & 0.3 \\
\end{bmatrix} \\
T_{100} &= T_{101} = \begin{bmatrix}
0.9 & 0.43333 & 0.13333 & 0.03333 \\
0.03333 & 0.5 & 0.33333 & 0.23333 \\
0.03333 & 0.03333 & 0.5 & 0.23333 \\
0.03333 & 0.03333 & 0.03333 & 0.5 \\
\end{bmatrix} \\
T_{010} &= T_{011} = \begin{bmatrix}
0.9 & 0.9 & 0.9 & 0.9 \\
0.03333 & 0.03333 & 0.03333 & 0.03333 \\
0.03333 & 0.03333 & 0.03333 & 0.03333 \\
0.03333 & 0.03333 & 0.03333 & 0.03333 \\
\end{bmatrix} \\
T_{110} &= T_{111} = \begin{bmatrix}
0.33333 & 0.03333 & 0.03333 & 0.03333 \\
0.6 & 0.33333 & 0.03333 & 0.03333 \\
0.03333 & 0.6 & 0.33333 & 0.03333 \\
0.03333 & 0.03333 & 0.6 & 0.9 \\
\end{bmatrix}
\end{align*}

\subsection{Setup of different runs}

\paragraph{Different Random Initial Distributions.}
In order to generate the results presented in \S~\ref{sec:sim_validation}, we solve the optimization problem (\ref{op:utilmaxeop}) once, using $\epsilon = 0.01$ and $c = 0.08$, and obtain the optimal policy $\maxuitleopstar$.
Subsequently, we perform simulations for 10 different random initial distributions $\initialdistribution$, where we observe the behavior of $\maxuitleopstar$ for the assumed dynamics \onesided\ over a duration of 200 steps.
In line with the Markov convergence theorem, all of these simulations yield the same stationary distribution. 
\paragraph{Different Dynamic Types.}
To obtain results in \S~\ref{sec:sim_validation}, we address two optimization problems: (\ref{op:utilmaxeop}) and (\ref{op:maxqual}). For each set of dynamics, we solve these problems independently, resulting in different values for $\maxuitleopstar$ and $\maxqualstar$ respectively.
Subsequently, we utilize the FICO distribution as the initial distribution $\initialdistribution$ and simulate the feature distribution for each policy over 200 steps, assuming the specified dynamics.

\subsection{Computational Resources and Run Time}

\paragraph{Computational Resources}
All experiments were conducted on a MacBook Pro (Apple M1 Max chip). Since we can efficiently solve the optimization problem, these experiments are executed on standard hardware, eliminating the necessity for using GPUs.

\paragraph{Run Time}
The optimization problems to find long-term policies in all experiments within this paper were consistently solved in under 10 seconds. Regarding the training of short-term fair policies on 5000 samples, the run times were approximately 20-23 minutes: {1245.92} seconds  for \texttt{short-EOP} ($\lambda=1$), {1244.25} seconds for \texttt{short-EOP} ($\lambda=2$), and { 1380.50} seconds for \texttt{short-MAXUTIL}.

%% file: sections/93_Appendix_Results-1.tex
\section{Additional Results}\label{apx:results}

In this section, we provide additional results related to the results discussed in \S~\ref{sec:simulations}. Our analysis centers around our guiding example, employing the data distributions sourced from FICO~\citep{fico2007} unless otherwise specified. The structure of this section is as follows:
\begin{itemize}
    \item In \S~\ref{apx:additional-start-distributions} we provide additional results for different starting distributions.
    \item In \S~\ref{apx:additional-static} we provide additional results for the comparison to short-term policies.
    \item In \S~\ref{apx:additional-epsilon} we provide additional results for varying the fairness threshold $\epsilon$ for our policy.
    \item In \S~\ref{apx:additional-types} we provide  additional results for the different dynamic types (\onesided, \recourse, \discouraged) that we introduced in the main paper.
    \item In \S~\ref{apx:additional-speed} we provide additional results for varying the speed at which feature changes occur (\slow, \medium, \fast).
    \item In \S~\ref{apx:additional-estimations} we provide  additional results for first sampling from FICO data and then estimating the distributions under partially observed labels.
\end{itemize}

\subsection{{Different Initial Starting Distributions}}\label{apx:additional-start-distributions}

\begin{figure}
\begin{center}
\includegraphics[width=\textwidth]{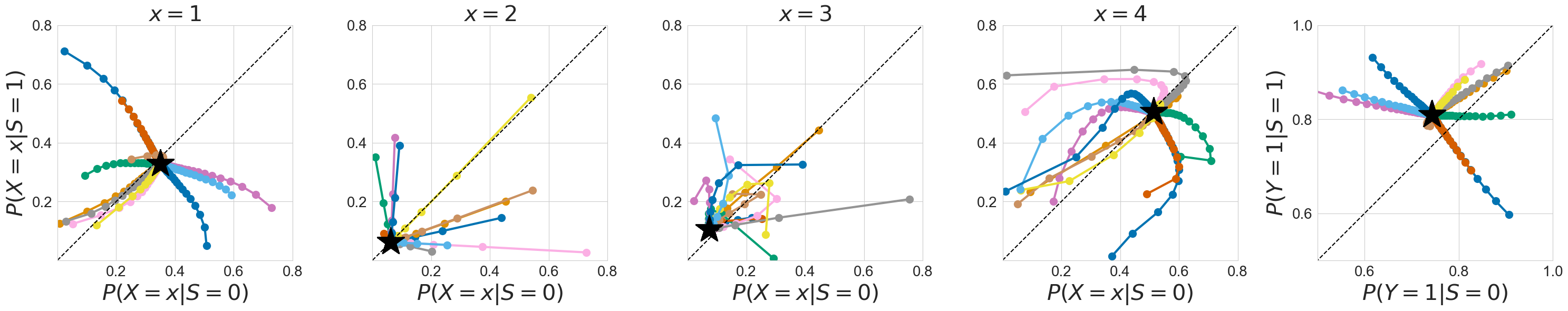}
\end{center}
\caption{Convergence of feature distributions for $\maxuitleopstar$ for different random starting distributions (colors) to unique stationary distributions $\stationary\!=\!\star$. Trajectories over $200$ time steps.  $c\!=\!0.8$, $\epsilon\!=\!0.01$.}
\label{fig:01-traj-Px-all-x}
\end{figure}

We provide additional results for the results shown \S~\ref{sec:sim_validation}, where we run simulations on 10 randomly sampled initial feature distributions $\initialdistribution$, setting $\epsilon=0.01, c=0.8$. In addition to the results shown in the main paper, we here display in Figure~\ref{fig:01-traj-Px-all-x} the resulting trajectories of all feature distributions.

%% file: sections/93_Appendix_Results-2.tex
\subsection{{Comparison to Static Policies}}\label{apx:additional-static}

We provide additional results comparing our long-term policy to short-term policies.

\paragraph{Static Policy Training.}
The short-term policies are logistic regression models implemented using PyTorch. The forward method computes the logistic sigmoid of a linear combination of the input features, while the prediction method applies a threshold of $0.5$ to the output probability to make binary predictions. The training process is carried out via gradient descent, with the train function optimizing a specified loss function. The \texttt{short-MAXUTIL} policy is trained using a binary cross-entropy loss. The fairness is enforced using a Lagrangian approach ($\lambda=2$). The \texttt{short-EOP} policy is trained using a binary cross-entropy loss and regularization terms measuring equal opportunity unfairness with $\lambda$ as hyperparameters controlling the trade-off between predictive accuracy and fairness. Training is performed for $2000$ epochs with a learning rate of $0.05$. We display results over $10$ random initializations. The experiments in the main paper are shown for \texttt{short-EOP} with $\lambda=2$. We show in the following results for different $\lambda$.

\begin{figure}
\begin{center}
\includegraphics[width=\textwidth]{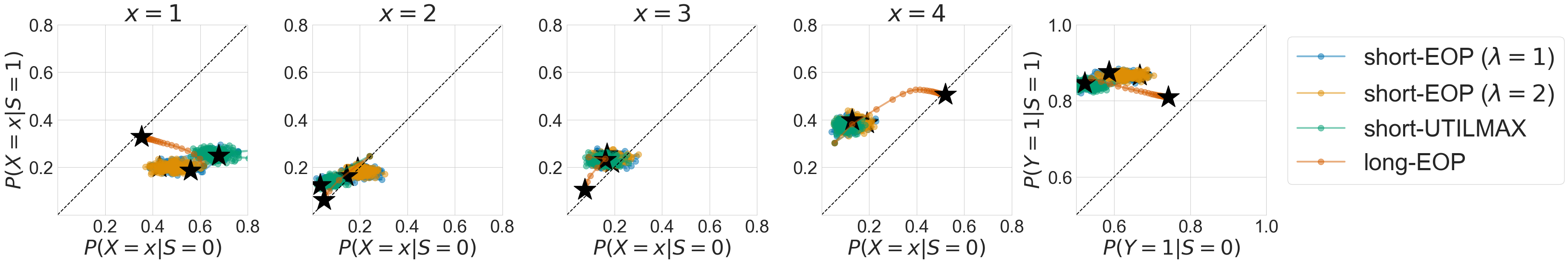}
\end{center}
\caption{Convergence of feature distributions for our long-term \texttt{long-EOP} ($\maxuitleopstar$) and the static policies (unfair: \texttt{short-MAXUTIL}, fair: \texttt{short-EOP} ($\lambda=2$). Trajectories over $200$ time steps. $c=0.8$, $\epsilon = 0.026$. Last distribution values are marked with $\star$.}
\label{fig:06-traj-Px-Py}
\end{figure}

\paragraph{Feature and Outcome Trajectories.}
Figure~\ref{fig:06-traj-Px-Py} presents the trajectories of our long-term \texttt{long-EOP} ($\maxuitleopstar$) and the static policies (unfair: \texttt{short-MAXUTIL}, fair: \texttt{short-EOP} ($\lambda=2$)) over 200 time steps for a single short-term policy seed.
We observe that our long-term policy converges to a stationary distribution and remains there once it has found it. In contrast, the trajectories of the short-term policies display non-stationarity, covering a wide range of distributions, as evidenced by the overlapping region. This indicates that the short-term policies exhibit a high variance and do not stabilize into a stationary distribution.

\begin{figure}
    \centering
    \includegraphics[width=0.9\textwidth]{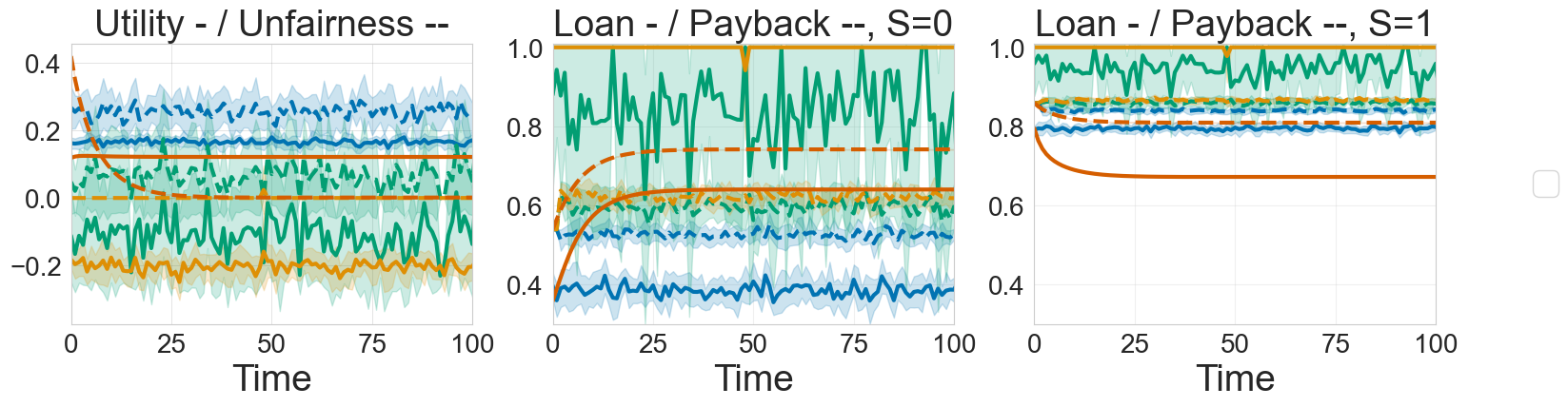}
    \caption{Results for our long-term \texttt{long-EOP} ($\maxuitleopstar$) and the static policies (unfair: \texttt{short-MAXUTIL}, fair: \texttt{short-EOP} ($\lambda=2$). . Top Left: Utility (solid, $\uparrow$) with $c=0.8$ and EOP-Unfairness (dashed, $\downarrow$). Top right / Bottom left: Loan (solid) and payback probability (dashed) per policy and sensitive $S$.}
    \label{fig:06-util-loan-apx}
\end{figure}

\paragraph{Utility, Fairness and Loan and Repayment Probabilities.}
Figure~\ref{fig:06-util-loan-apx} (top left) displays $\utility$ and $\eopunf$ over the first 100 time steps.
We observe that short-term policies, which are updated at each time step, tend to exhibit greater variance compared to the long-term policy, which remains fixed at $t=0$ - even as the underlying data distribution evolves in response to decision-making. Among the two short-term fair policies, the fairer one ($\lambda=2$) approaches nearly zero unfairness, whereas the less fair one ($\lambda=1$) displays a higher level of unfairness. Specifically, the more fair policy ($\lambda=2$) reaches a low (negative) utility, while the less fair one ($\lambda=1$) maintains a higher (though still negative) utility.
The unfair short-term policy (\texttt{UTILMAX}) achieves positive utility but does so at the cost of a high level of unfairness. This highlights the trade-off between fairness and utility that short-term policies encounter.
Conversely, our long-term fair policy maintains a level of unfairness close to zero while experiencing only a modest reduction in utility compared to the unfair short-term policy. This underscores our policy's capacity to attain long-term fairness while ensuring a higher level of utility, leveraging the long-term perspective to effectively shape the population distribution.

Figure~\ref{fig:06-util-loan-apx} (top right, bottom left) presents the loan probability $\Pbb(D=1\mid S=s)$ and payback probability $\Pbb(Y=1\mid S=s)$ for non-privileged ($S=0$) and privileged ($S=1$) groups.
In addition to the results presented in the main paper (Figure~\ref{fig:06-util-loan}), we observe a difference between the two short-term fair policies in our analysis in this appendix. The more equitable policy ($\lambda=2$) achieves a low level of unfairness by granting loans with a probability of 1 to individuals across all social groups. The less equitable policy ($\lambda=1$) provides loans to the underprivileged group with an average probability of approximately 0.85, while the privileged group receives loans at an average probability of around 0.9.

Crucially, the less equitable policy ($\lambda=1$) exhibits a much higher variability in loan approval probabilities for the underprivileged group across different time steps compared to the privileged group. This highlights that unfairness does not solely manifest at the mean level but also in the variability across time. Both policies tend to grant loans at probabilities exceeding the actual repayment probabilities within the population. This suggests an "over-serving" phenomenon, implying that the policies on average extend loans to individuals who may not meet the necessary qualifications for borrowing.

In contrast, our policy maintains stability and converges to a low difference in loan approval probabilities between groups without significant temporal variance. Importantly, our loan approval probabilities remain below the loan repayment (as for the short-term unfair policy (\texttt{UTILMAX})) probabilities, indicating that, on average, the policies are extending loans to individuals who are indeed eligible for them. In addition, for our policy, the gap between loan provision and repayment probabilities is similar across sensitive groups.

\begin{figure}
      \centering
    \includegraphics[width=0.95\textwidth]{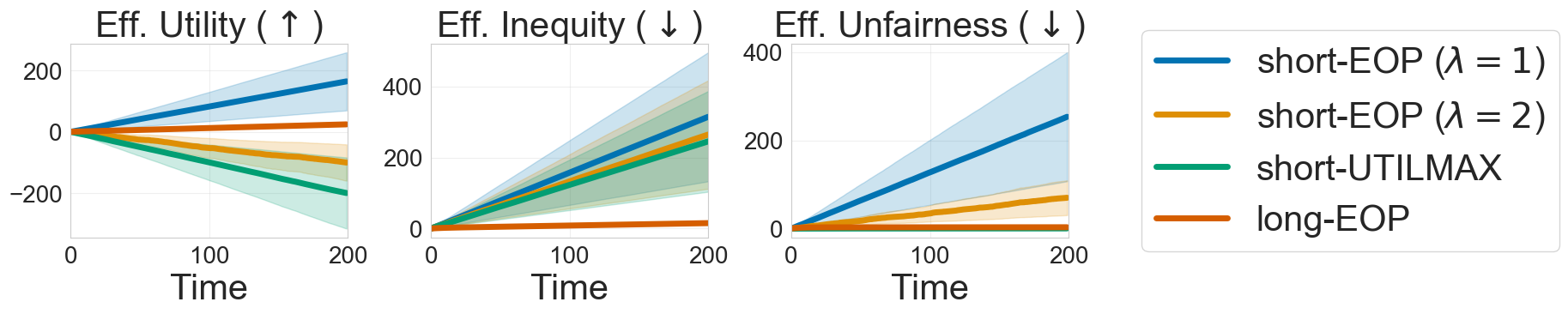}
 \caption{Results for our long-term \texttt{long-EOP} ($\maxuitleopstar$) and the static policies (unfair: \texttt{short-MAXUTIL}, fair: \texttt{short-EOP} ($\lambda=2$). Effective (cumulative) utility $\utility$, inequity $\inequity$, and (EOP) unfairness $\eopunf$ for different policies.}\label{fig:06-eff-measures}
\end{figure}

\paragraph{Effective Utility, Inequity and Unfairness.}
Figure \ref{fig:06-eff-measures} illustrates effective (accumulated) measures of utility, inequity, and (EOP) unfairness over time for the different policies, where results for static policies are reported over $10$ random initializations.
We observe that the short-term unfair policy (\texttt{short-UITLMAX} consistently accumulates the highest utility across all dynamics, while simultaneously maintaining a high level of effective unfairness and inequality. Conversely, the short-term fair policies ($\texttt{short-EOP} (\lambda=1$) and $(\lambda=2)$) exhibit negative effective utility, but they do achieve lower levels of effective fairness and inequity.

For our long-term policy (\texttt{long-EOP}), we find that it accumulates positive utility over time. Although its utility remains below that of the short-term unfair policy, our policy exhibits very low levels of effective unfairness. Importantly, it also yields minimal accumulated inequity, even though it was not specifically optimized for this.

Analyzing the cumulative effects of policies is essential for evaluating the long-term impact of each policy choice. This analysis can, for instance, help determine whether investing in fairness pays off in the long-term and whether sacrificing short-term fairness in the initial stages ultimately benefits society in the long run.

%% file: sections/93_Appendix_Results-3.tex
\subsection{Different Fairness Levels}\label{apx:additional-epsilon}
We provide additional results, where we use the initial distribution $\initialdistribution$ from FICO and solve the optimization problem (\ref{op:utilmaxeop}) for four different fairness levels
$\epsilon$. This results in four policies $\maxuitleopstar$.

\begin{figure}
\begin{center}
\includegraphics[width=\textwidth]{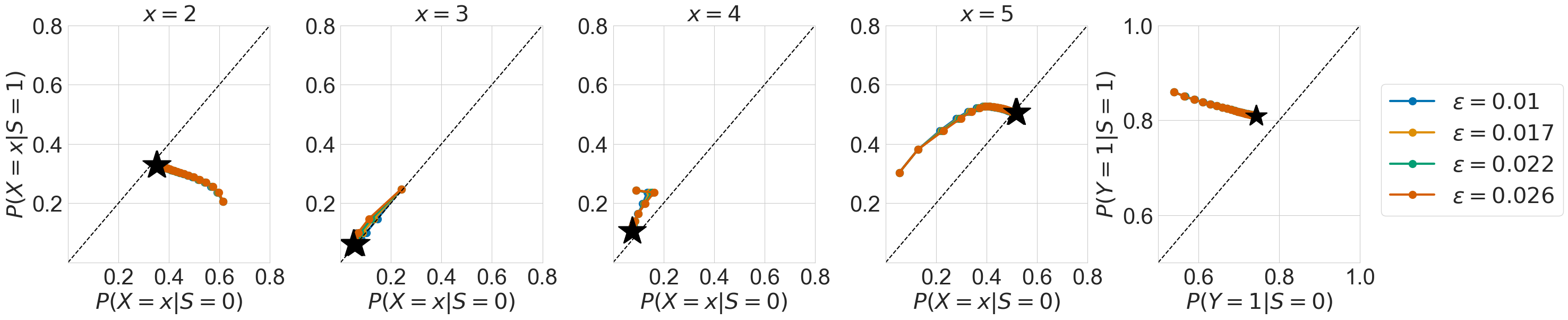}
\end{center}
\caption{Convergence of feature distributions for $\maxuitleopstar$ for different fairness thresholds $\epsilon$ to unique stationary distributions $\stationary=\star$. Trajectories over $200$ time steps. $c=0.8$.}
\label{fig:02-traj-Px-Py}
\end{figure}

\paragraph{Feature and Outcome Trajectories.}
Figure~\ref{fig:02-traj-Px-Py} presents the trajectories of $\maxuitleopstar$ over 200 time steps for different fairness thresholds $\epsilon$. We observe that although the convergence process, time, and final stationary distribution ($\star$) are very similar for different targeted fairness levels.

\begin{figure}
    \centering
    \includegraphics[width=0.9\textwidth]{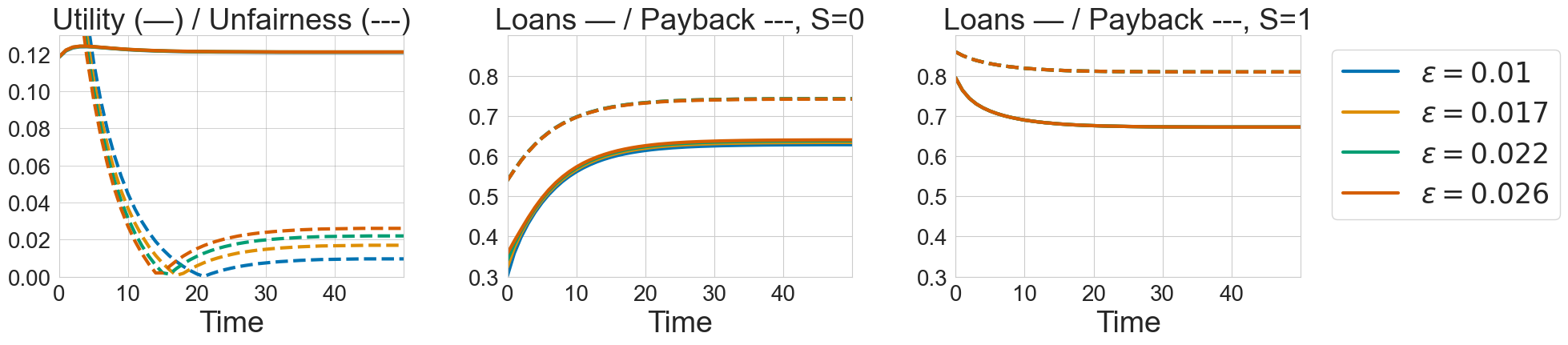}
    \caption{Results for different $\epsilon$-EOP-fair $\maxuitleopstar$. Top Left: Utility (solid, $\uparrow$) with $c=0.8$ and EOP-Unfairness (dashed, $\downarrow$). Top right / Bottom left: Loan (solid) and payback probability (dashed) per policy and sensitive $S$.}
    \label{fig:02-util-loan}
\end{figure}

\paragraph{Utility and Loan and Repayment Probabilities.}
Figure~\ref{fig:02-util-loan} (top left) displays $\utility$ and $\eopunf$ over the first 50 time steps (until convergence). 
We observe that all policies converge to a similar utility level while maintaining their respective $\epsilon$ level, confirming the effectiveness of our optimization problem.
Figure~\ref{fig:02-util-loan} (top right, bottom left) presents the loan probability $\Pbb(D=1\mid S=s)$ and payback probability $\Pbb(Y=1\mid S=s)$ for non-privileged ($S=0$) and privileged ($S=1$) groups. 
While the probabilities across sensitive groups ultimately stabilize close together in the long term, the initial 20 steps exhibit a large difference in loan and payback probabilities.
Optimizing for long-term goals may thus lead to unfairness in the short term, and it is important to carefully evaluate the potential impact of this on public trust in the policy.

%% file: sections/93_Appendix_Results-4.tex
\subsection{Different Dynamic Types}\label{apx:additional-types}

Results in this subsection are for different dynamic types: \onesided, \recourse, and \discouraged. See~\ref{apx:dynamics} for more details on these specific dynamics. We solve both optimization problems for each of the three dynamics, where solving (\ref{op:utilmaxeop}) provides $\maxuitleopstar$ and solving (\ref{op:maxqual}) provides $\maxqualstar$. 

\begin{figure}
\begin{center}
\includegraphics[width=\textwidth]{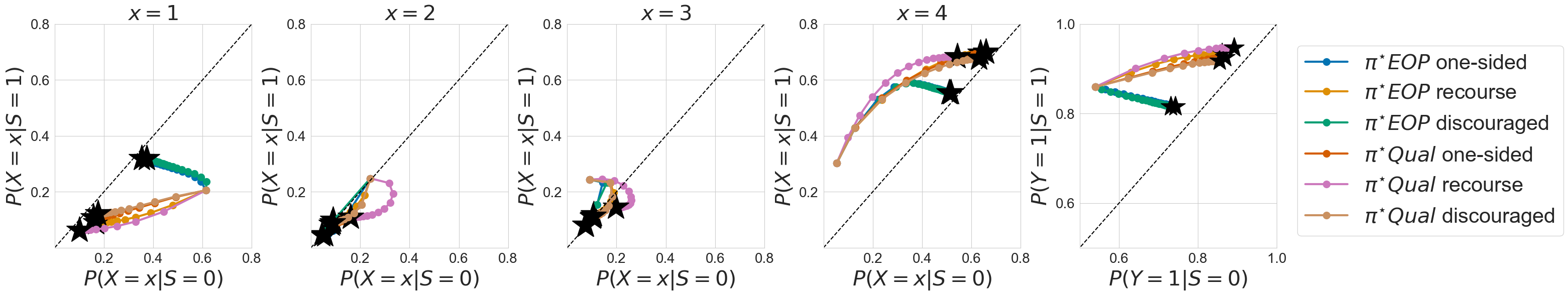}
\end{center}
\caption{{Convergence of $\maxuitleopstar$ and $\maxqualstar$ for different type of dynamics towards different unique stationary distributions $\stationary=\star$. Trajectories over $200$ time steps. Top four plots: feature distribution $\stationaryt$. Bottom left: distribution of the outcome of interest. Equal feature/outcome distribution dashed. Initial distribution $\initialdist=$FICO, $c=0.8$, $\epsilon = 0.01$.}}
\label{fig:04-traj-Px-Py}
\end{figure}

\paragraph{Feature and Outcome Trajectories.}
Figure~\ref{fig:04-traj-Px-Py} presents the trajectories of $\maxuitleopstar$ and $\maxqualstar$ over 200 time steps for different types of dynamics. We observe that although the initial distribution remains unchanged, the convergence process, time, and final stationary distribution ($\star$) differ depending on the dynamics. Notably, the stationary distribution of $\maxqualstar$ appears to be similar for \onesided\ and \discouraged\ dynamics. On the other hand, the results for all other dynamics and policies demonstrate distinct but relatively close outcomes.

\begin{figure}
\begin{center}
\includegraphics[width=0.9\textwidth]{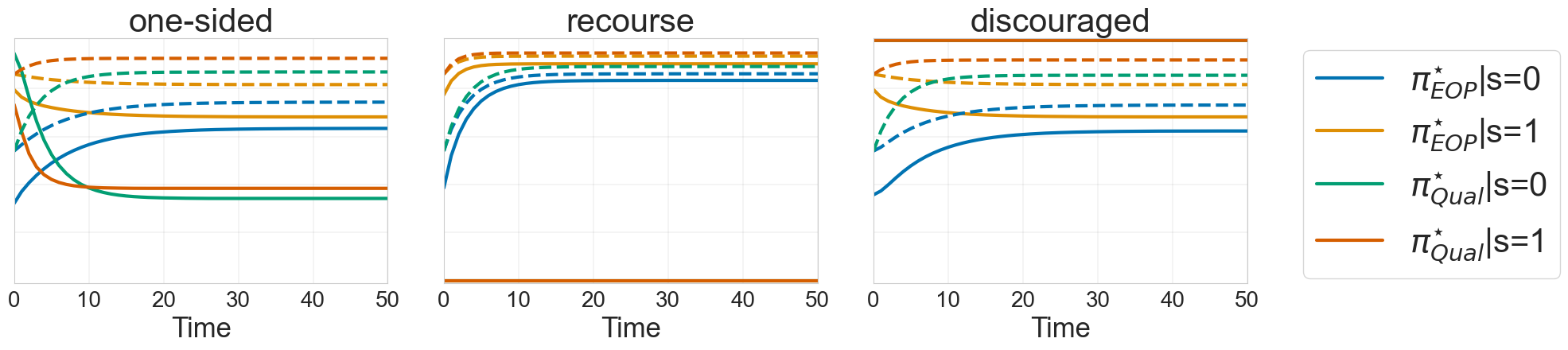}
\end{center}
\caption{Loan probability $\Pbb(D=1\mid S=s)$ (solid) and repayment probability $\Pbb(Y=1 \mid S=s)$ (dashed) for different type of dynamics (one-sided, recourse, discouraged) and policies $\maxuitleopstar, \maxqualstar$ per sensitive attribute $s\in \{0,1\}$. Initial distribution $\initialdist=$ FICO, $c=0.8$, $\epsilon = 0.01$.}
\label{fig:04-util-loan}
\end{figure}

\paragraph{Utility, Fairness and Loan and Repayment Probabilities.}
Figure \ref{fig:04-util-loan} showcases the group-dependent probabilities of receiving a loan, $\Pbb(D_t=1\mid S=s)$, and repayment, $\Pbb(Y_t=1\mid S=s)$, for both the non-privileged ($S=0$) and privileged ($S=1$) groups. 
The probabilities are displayed for the convergence phase (first 50 time steps) for policies $\maxuitleopstar$ and $\maxqualstar$ across dynamics types.
When the payback probabilities are higher compared to the loan probabilities, it suggests an underserved community where fewer credits are granted than would be repaid. 
In the case of \onesided\ dynamics, we find that for $\maxuitleopstar$, the loan and repayment probabilities are relatively close to each other at each time step. However, for $\maxqualstar$, the gap between repayment and loan probabilities widens as time progresses. At convergence, both sensitive groups exhibit a repayment rate of approximately 0.8, while the loan-granting probability is around 0.4.
This suggests that, in the \onesided\ dynamics, for $\maxqualstar$ the repayment rate is higher compared to the loan granting rate, indicating that a significant number of individuals who would repay their loan are not being granted one.
In the case of \onesided\ dynamics, similar to the \discouraged\ dynamics, we observe different short-term and long-term effects. Specifically, for $\maxuitleopstar$, the probability of receiving a loan initially differs between the sensitive groups within the first 20 time steps. However, as time progresses, these probabilities tend to become closer to each other. This suggests a potential reduction in the disparity of loan access between the sensitive groups over time under the influence of the $\maxuitleopstar$ policy.
In the case of \recourse\ dynamics, we observe that the loan granting and repayment probabilities tend to stabilize closely together in the long term across sensitive groups and under both policies---except for $\maxqualstar$ when $S=1$. In this particular case, the $\maxqualstar$ policy sets $\policy(D=1 \mid X=x, S=1) = 0$ for all values of $x$. This scenario serves as an example where optimizing for long-term distributional goals without enforcing predictive fairness constraints can lead to individuals with a high probability of repayment being consistently denied loans. 
%

%% file: sections/93_Appendix_Results-5.tex
\subsection{Different Dynamic Speeds}\label{apx:additional-speed}
We begin by assuming one-sided dynamics and then introduce variation in the speed of transitioning between different credit classes. This variation encompasses three levels: \slow, \medium, and \fast, each representing the rate at which borrowers' credit scores evolve in response to decisions. Additional information about these specific dynamics can be found in Section~\ref{apx:dynamics}.
For each of these three dynamics, we address both optimization problems. Solving (\ref{op:utilmaxeop}) yields $\maxuitleopstar$, while solving (\ref{op:maxqual}) provides $\maxqualstar$." 

\begin{figure}
\begin{center}
\includegraphics[width=\textwidth]{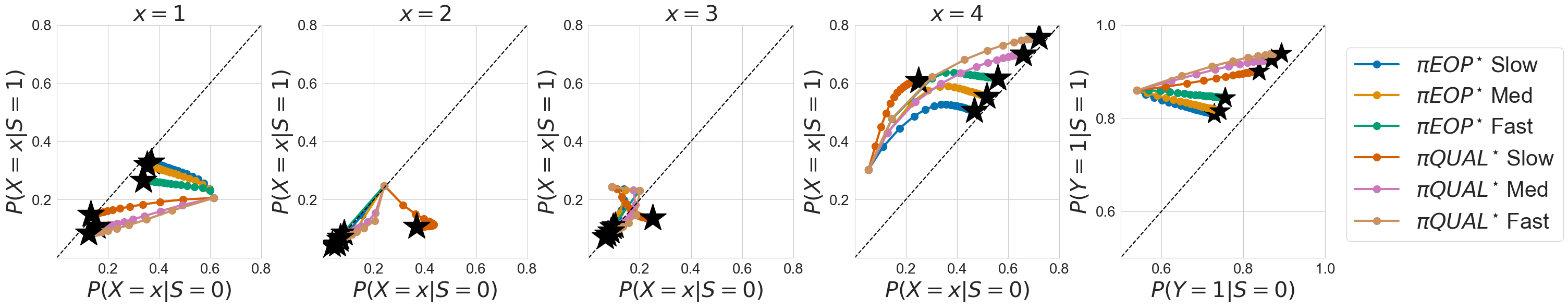}
\end{center}
\caption{{Convergence of $\maxuitleopstar$ and $\maxqualstar$ for different speeds of dynamics towards different unique stationary distributions $\stationary=\star$. Trajectories over $200$ time steps. Left four plots: feature distribution $\stationaryt$. Right: distribution of the outcome of interest. Equal feature/outcome distribution dashed. Initial distribution $\initialdist=$FICO, $c=0.8$, $\epsilon = 0.01$.}
}
\label{fig:03-traj-Px-Py}
\end{figure}

\paragraph{Feature and Outcome Trajectories.}
Figure \ref{fig:03-traj-Px-Py} depicts the trajectories over 200 time steps for $\maxuitleopstar$ and $\maxqualstar$ under different speeds of one-sided dynamics. While the initial distribution remains the same for all runs, the convergence process, time, and final stationary distribution ($\star$) vary depending on the dynamics speed. 
Regarding the group-dependent distribution of $Y$, we observe that $\maxqualstar$ achieves a higher distribution (which in addition is closer to the equal outcome distribution) compared to $\maxuitleopstar$. This can be attributed to the fact that $\maxqualstar$ explicitly optimizes for maximizing the total distribution of $Y$. Additionally, we notice that for both policies slower dynamics result in lower stationary distributions of $Y$ compared to faster dynamics.

\begin{figure}
      \centering
    \includegraphics[width=0.9\textwidth]{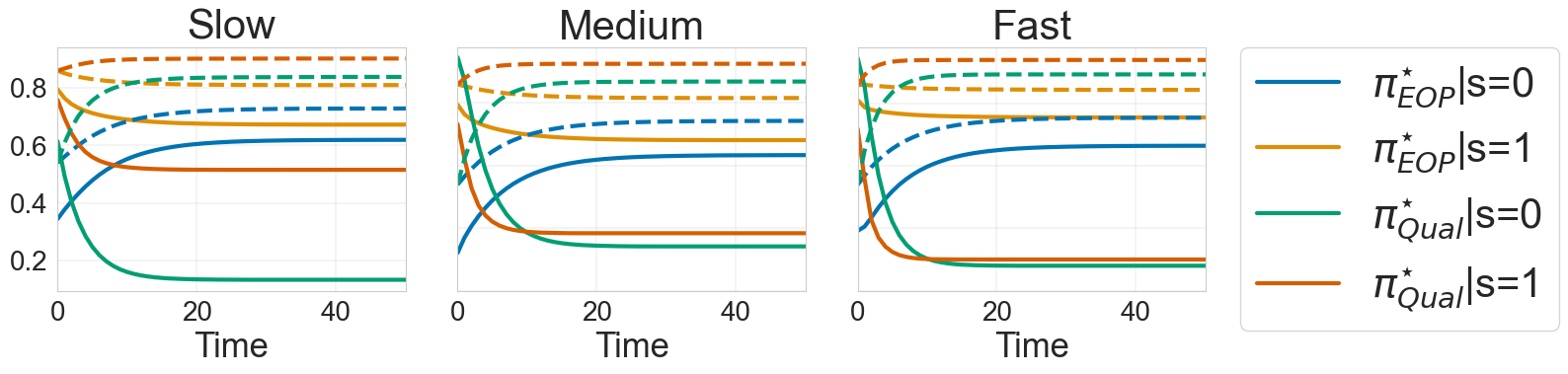}
 \caption{Loan probability $\Pbb(D=1\mid S=s)$ (solid) and repayment probability $\Pbb(Y=1 \mid S=s)$ (dashed) for different speed of one-sided dynamics (\slow, \medium, \fast) and policies $\maxuitleopstar, \maxqualstar$ per sensitive attribute $s\in \{0,1\}$. Initial distribution $\initialdist=$ FICO, $c=0.8$, $\epsilon = 0.01$.}
         \label{fig:03-util-loan}
\end{figure}

\paragraph{Utility, Fairness and Loan and Repayment Probabilities.}
Figure~\ref{fig:03-util-loan} depicts the group-dependent probabilities of receiving a loan, $\Pbb(D=1\mid S=s)$, and repayment, $\Pbb(Y=1\mid S=s)$, for both non-privileged ($S=0$) and privileged ($S=1$) groups. 
The probabilities are shown for the convergence phase (initial 50 time steps) of policies $\maxuitleopstar$ and $\maxqualstar$ across different speeds of one-sided dynamics. 
Higher payback probabilities compared to loan probabilities can indicate an underserved community where fewer credits are granted than would be repaid. 
Across all dynamics, we observe small differences in the repayment distributions for each policy. The repayment probabilities are consistently higher for the non-protected group compared to the protected group. Moreover, in general, $\maxqualstar$ yields higher repayment rates than $\maxuitleopstar$.
However, the loan probabilities---which indicate a group's access to credit---exhibit differences across dynamics and policies. As expected, the utility-maximizing $\maxuitleopstar$ generally provides higher loan rates compared to $\maxqualstar$. While the loan rates remain similar across dynamics for $\maxuitleopstar$, they vary for $\maxqualstar$.
Under \slow\ dynamics, $\maxqualstar$ yields low loan probabilities for the protected group, which then increases for \medium\ and \fast\ dynamics. 
Furthermore for $\maxqualstar$, the discrepancy between acceptance rates for sensitive groups is greatest at \slow\ dynamics, and decreases significantly at \medium\ dynamics - at the expense of the non-protected group. Finally, for \fast\ dynamics, the acceptance rates for sensitive groups are approximately equal.

These observations emphasize the importance of conducting further investigations into the formulation of long-term goals, taking into account their dependence on dynamics and the short-term consequences. This includes not only considering the type of dynamics (one-sided or two-sided), but also the speed at which individuals' feature changes in response to a decision.

\begin{figure}
      \centering
    \includegraphics[width=0.9\textwidth]{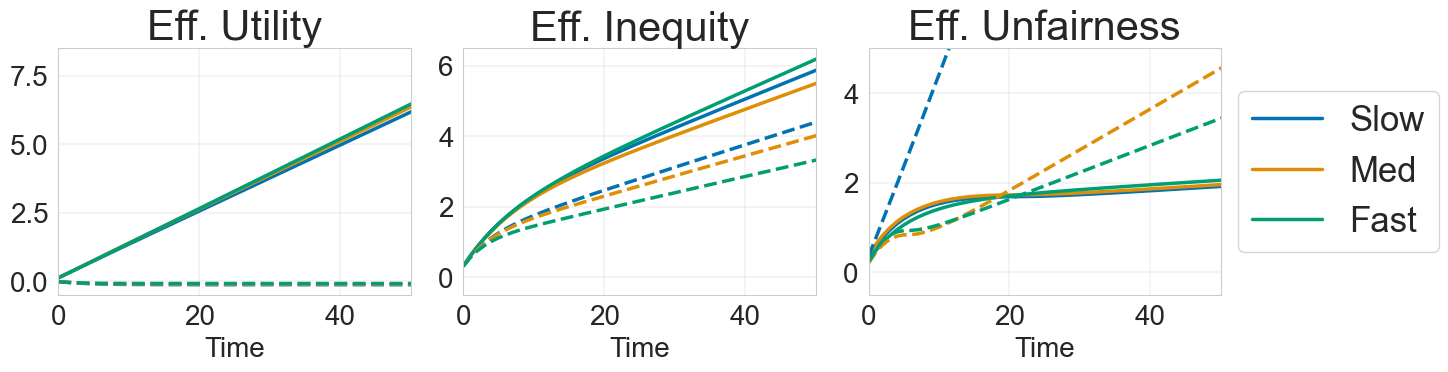}
 \caption{Effective (cumulative) utility $\utility$, inequity $\inequity$, and (EOP) unfairness $\eopunf$ for different policies ($\maxuitleopstar$ solid, $\maxqualstar$ dashed). }
         \label{fig:03-eff-measures}
\end{figure}

\paragraph{Effective Utility, Inequity and Unfairness.}
Figure \ref{fig:03-eff-measures} illustrates effective (accumulated) measures of utility, inequity, and (EOP) unfairness over time.
For all dynamics, the policies align with their respective targets. $\maxuitleopstar$ accumulates the highest utility across all dynamics while maintaining a low effective unfairness after an initial convergence period. On the other hand, $\maxqualstar$ exhibits a small negative effective utility due to the imposed zero-utility constraint, but achieves lower effective inequity by maximizing the total distribution of the outcome of interest.
We observe that the speed of dynamics does not significantly affect effective utility for both policies and effective unfairness for the $\maxuitleopstar$ policy. 
However the speed of dynamics does have an impact effective inequity, although its effect varies for each policy.
Among the $\maxuitleopstar$ policies, we find that the \medium\ dynamics result in the lowest effective inequity, whereas among the $\maxqualstar$ policies, the \fast\ dynamics exhibit the lowest effective inequity.
While the effective utility is minimally affected by the speed of dynamics in the case of $\maxuitleopstar$, we observe different results for effective inequity.
Among the $\maxuitleopstar$ policies, the \medium\ dynamics result in the lowest effective inequity. Conversely, among the $\maxqualstar$ policies, the \fast\ dynamics exhibit the lowest effective inequity.
These observations highlight that the final outcomes of decision policies are not only influenced by the type of dynamics (one-sided and two-sided), but also by the speed of dynamics. It is thus crucial to also consider the rate at which individuals are able to change features within one time step. 
This consideration can for example be important in the context of recourse, where not all individuals may have the ability to implement the minimum recommended actions, potentially due to individual limitations. Consequently, only a fraction of individuals would be able to move up in their credit class in response to a negative decision. 

%% file: sections/93_Appendix_Results-6.tex
\subsection{Dynamics Estimation under Partially Observed Labels}
\label{apx:additional-estimations}

We conduct additional experiments to investigate the impact of estimation errors in the underlying distributions on the quality of results.
In a more realistic loan example, label $Y$ might be partially observed (i.e., observed only for individuals who received a positive loan decision). In this case, the estimate of $Y$ may no longer be as accurate for one sensitive group as for another. We investigate the sensitivity of our derived policy to the estimation of $Y$ for different decision policies (which reveal different amounts of labels for different subgroups) compared to access to the true distribution of $Y$.
We first generate a temporal dataset comprising two time steps. These samples were drawn from the FICO base distribution, and we assumed the dynamics of {One-sided General} (as described in \S~\ref{apx:dynamics}). The dataset is comprised of 50,000 samples aligning with the dataset scales employed in the fairness literature, such as the Adult dataset~\cite{AdultData}.
We deploy three different policies that influence the data observed at $t=1$, \texttt{random}, \texttt{threshold}, \texttt{biased}, with the following formulations:
\begin{itemize}
    \item \randompol\ is defined by $\Pbb(D=1 \mid X, S) = 0.5$ for all $X, S$;
    \item \biaspol\ is defined for all $S$ by $\Pbb(D=1 \mid X, S) = 0.1$ if $X<=2$ and for $S=0$ as $\Pbb(D=1 \mid X, S) = 0.3$ if $X>2$ and for $S=1$ as $\Pbb(D=1 \mid X, S) = 0.9$. 
\end{itemize}

\begin{figure}
    \centering
    \begin{subfigure}{0.32\textwidth}
        \includegraphics[width=\textwidth]{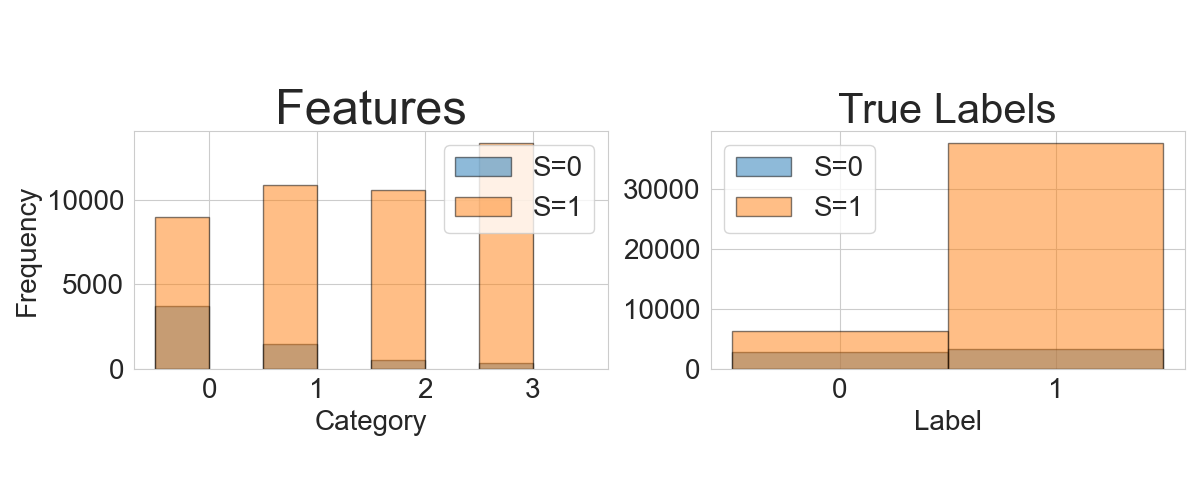}
        \caption{True distributions of features and labels.}
        \label{fig:dist-true}
    \end{subfigure}
    \hfill
    \begin{subfigure}{0.32\textwidth}
        \includegraphics[width=\textwidth]{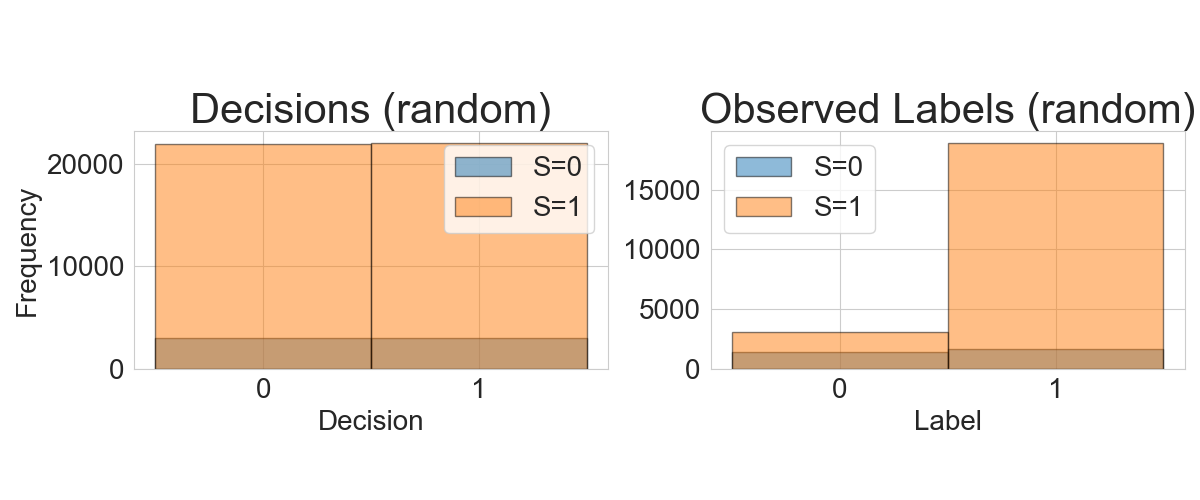}
        \caption{Distribution of decisions and observed labels for \randompol.}
        \label{fig:dist-randompol}
    \end{subfigure}
    \hfill
    \begin{subfigure}{0.32\textwidth}
        \includegraphics[width=\textwidth]{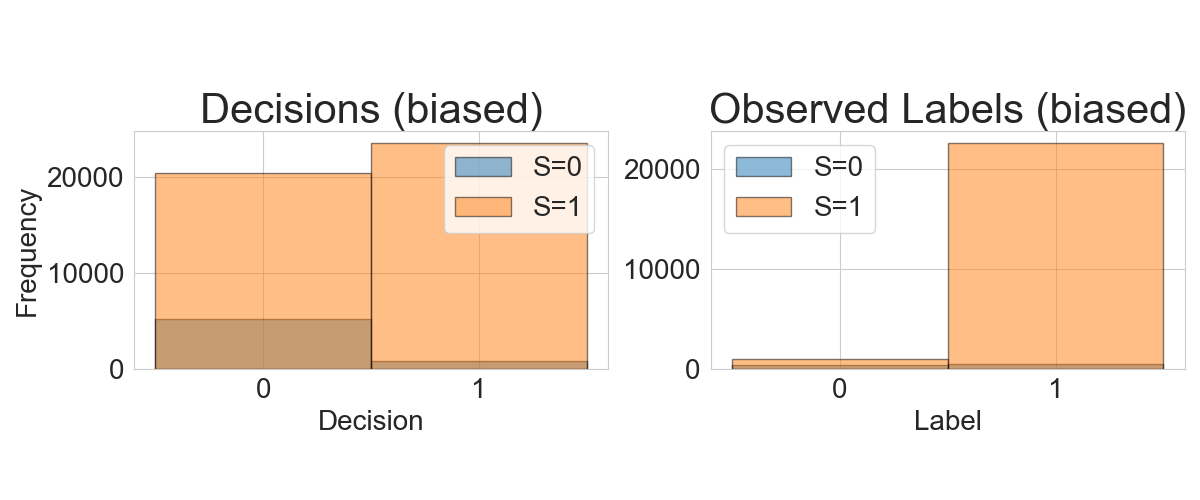}
        \caption{Distribution of decisions and observed labels for \biaspol.}
        \label{fig:dist-biaspol}
    \end{subfigure}
    \caption{Data distributions for different temporal datasets based on FICO used to estimate label distributions and dynamics.}
    \label{fig:combined-figures}
\end{figure}

The true distribution of features and label at $t=0$ are shown in Figure~\ref{fig:dist-true}. The distributions of decisions and observed labels under the different policies are shown in Figures~\ref{fig:dist-randompol} - \ref{fig:dist-biaspol}.

We then estimate both $\labelfunctall$ and $\dynamicfunctall$ from the observed samples, with the latter being dependent on the former. Subsequently, we solve the optimization problem ($c=0.9$, $\epsilon = 0.00005$) using these estimated distributions yielding three different policies (one per estimation). Consequently, we simulate the performance of the discovered policies under the true distributions and $\initialdist=$FICO. In the evaluation, we compare the results to the policy obtained under the true probability estimate $\labelfunctall$ as supplied by FICO (\truepol).

\begin{figure}
\begin{center}
\includegraphics[width=\textwidth]{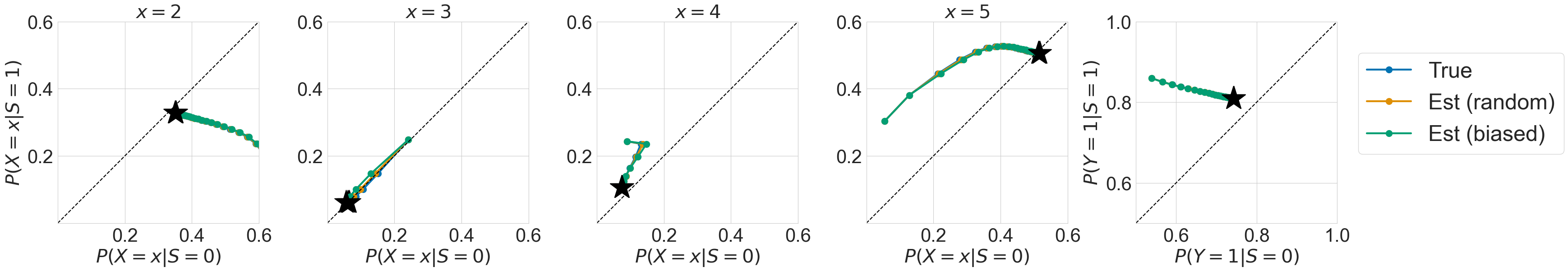}
\end{center}
\caption{Convergence of $\maxuitleopstar$ under true and estimations of $\labelfunctall$ and $\dynamicfunctall$ and under different type of initial policies (\randompol, \threspol, \biaspol). $200$ time steps, last time step marked $\star$. Top four plots: feature distribution $\stationaryt$. Bottom left: distribution of the outcome of interest. Equal feature/outcome distribution dashed.}
\label{fig:05-traj-Px-Py}
\end{figure}

\paragraph{Feature and Outcome Trajectories.}
Figure~\ref{fig:05-traj-Px-Py} displays the trajectories of $\maxuitleopstar$ for 200 time steps for the optimal policies obtained under both the true and estimated distributions and dynamics. Notably, the initial distribution remains the same, and the policies slightly vary in their convergence process to the stationary distribution ($\star$), while staying close to each other. 
It is important to emphasize that all policies successfully achieve a stationary distribution. This is due to the fact that even though we employ estimated distributions as inputs for the optimization problem, we are still solving the optimization problem for a policy that induces a stationary distribution that satisfies the fairness criteria. We showcase this in the next results.

\begin{figure}
    \centering
    \includegraphics[width=0.9\textwidth]{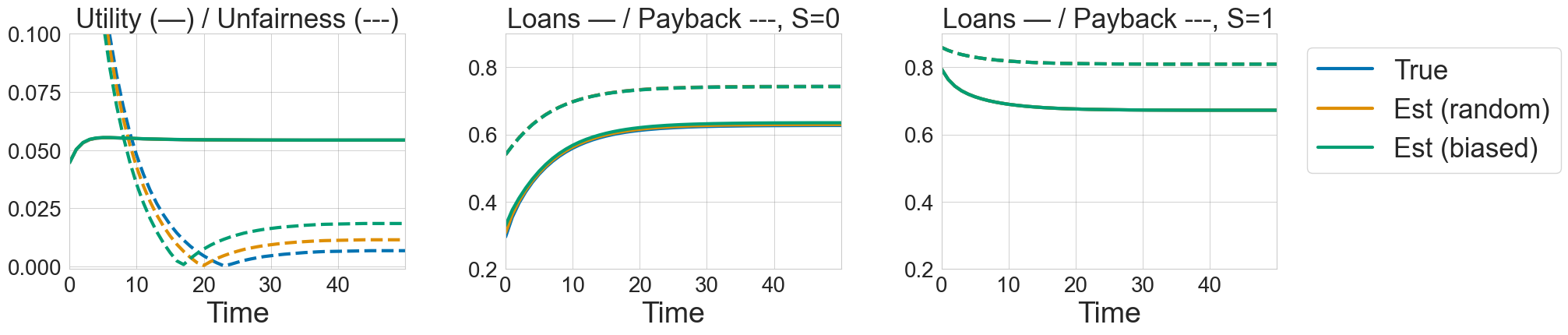}
    \caption{Results for our $\maxuitleopstar$ under true and estimations of $\labelfunctall$ under different type of initial policies (\randompol, \threspol, \biaspol). Top Left: Utility (solid, $\uparrow$) and EOP-Unfairness (dashed, $\downarrow$) over first $50$ time steps. 
    Remaining: Loan (solid) and payback probability (dashed) per policy and sensitive $S$. }
    \label{fig:05-util-loan}
\end{figure}

\paragraph{Utility, Fairness and Loan and Repayment Probabilities.}
Figure~\ref{fig:05-util-loan} (left) displays $\utility$ and $\eopunf$ over the first 50 time steps (until convergence).
We observe that the policies exhibit a different level of unfairness, while still achieving low unfairness.
The policy derived from the true probabilities and dynamics achieves lowest unfairness, the policy derived from probabilities and dynamics collected under a random policy has slightly higher unfairness, and the policy derived from probabilities and dynamics collected under a biased policy has the highest unfairness.
In terms of utility, where we aim for maximization without imposing a strict constraint, we observe that all policies exhibit a similar utility level.
Figure~\ref{fig:02-util-loan} (middle, right) displays the loan probability $\Pbb(D=1\mid S=s)$ and payback probability $\Pbb(Y=1\mid S=s)$ for non-privileged ($S=0$) and privileged ($S=1$) groups.
While there is no difference in loan and payback probabilities for the privileged group ($S=1$) between the policies, we observe a small difference for the unprivileged group ($S=0$). The policy derived from true probabilities and dynamics provides fewer loans to the unprivileged group compared to the policy derived from probabilities and dynamics collected under the random policy. Interestingly, the policy derived from probabilities and dynamics collected under a biased policy grants the most loans to the unprivileged group. Note, that our unfairness metric in the left plot is equal opportunity~\cite{hardt2016equality}, not demographic parity~\cite{dwork2012fairness}. Consequently, this observation may be explained by the policy obtained from biased estimation providing loans to a higher number of individuals from the unprivileged group who may not be able to repay them.
Thus, while we do achieve a stationary distribution using estimated probabilities, it is important to note that convergence to the intended fair state is not guaranteed when estimation errors are present. 
However, if the estimations closely approximate the true distribution, the resulting stationary distribution achieves similar utility and fairness properties as the stationary distribution that would have been achieved had the policy found under the true probabilities.
%

%% file: sections/94_Appendix_Guiding-example.tex
\section{Example Scenarios}\label{apx:example}

\subsection{Assumptions of the Guiding Example}
%
In this section, we discuss the assumptions taken in the data generative model introduced in \S~\ref{sec:example}. 

\begin{Asm}\label{ass:s-root}
$S$ is a root node and $X_t$, $Y_t$ and $D_t$ (potentially) depend on $S$.
\end{Asm}

It is commonly assumed in the causality and fairness literature that sensitive features are root nodes in the graphical representation of the data generative model~\citep{kusner2017counterfactual, chiappa2019path, kilbertus2020sensitivity}, although there is some debate on this topic~\citep{mhasawade2021causal, hu2020s}.
The assumption that the sensitive attribute $S$ influences $X_t$ is based on the observation that in practical scenarios, nearly every (human) characteristic is causally influenced by the sensitive attribute~\citep{ kusner2017counterfactual, chiappa2019path}.
In some cases, it is also assumed that $S$ influences $Y_t$ \citep{chiappa2019path}, while in other cases, this assumption is not made \citep{liu2018delayed}.
%
The extent to which the decision $D_t$ is directly influenced by the sensitive attribute $S$ depends on the decision policy being employed.
Policies that strive for (statistical) fairness often require explicit consideration of the protected attribute in their decision-making process~\citep{hardt2016equality, dwork2012fairness, corbett2017algorithmic}.

\begin{Asm}\label{ass:x-to-y}
The outcome of interest $Y_t$ depends on features $X_t$.
\end{Asm}

The assumption that changes in $X_t$ lead to changes in $Y_t$ is prevalent in scenarios involving lending~\citep{liu2018delayed, creager2020causal, damour2020fairness, Hu_Zhang_2022}.
This assumption is also implicit in problems where individuals seek recourse, e.g., via minimal consequential recommendations \citep{karimi2021survey} or social learning \citep{heidari2019effort}.

 \begin{Asm}\label{ass:x-to-d}
    Decision $D_t$ depends on features $X_t$.
\end{Asm}
In algorithmic decision-making, the primary objective of a policy is typically to predict the unobserved label or outcome of interest, denoted as $Y$, based on the observable features, denoted as $X$~\citep{scholkopf2012causal}.
We make the assumption that an individual's observed features at a particular point in time are sufficient to make a decision and conditioned on these features, the decision is independent of past features, labels, and decisions. This assumption aligns with prior work in the field~\citep{zhang2020fair, creager2020causal, karimi2021survey}.

\begin{Asm}\label{asm:s-immutable}
An individual's sensitive attribute $S$ is immutable over time.
\end{Asm}
For simplicity, we assume that individuals do not change their sensitive attribute. This assumption aligns with previous works that consider a closed population~\citep{liu2018delayed, creager2020causal, damour2020fairness, von2020fairness}. A closed population refers to a group of individuals that remains constant throughout the study or analysis. It implies that there are no additions or removals from the population of interest.
Other work considers that individuals join and leave the population over time, leading to a changing distribution of the sensitive attribute~\citep{hashimoto2018fairness}. The assumption that individuals do not change their sensitive attribute is controversial because, on the one hand, social categories are often ontologically unstable~\citep{fairmlbook, hu2020s}, and as such their boundaries are not clearly defined and dynamic. On the other hand, it ignores that individuals may be assigned identities at birth which they have the agency to correct at a given time. For example, an individual assigned one religion at birth may have a different religion at a later stage in life. 
\begin{Asm}
An individual's next step's features $X_{t+1}$ depend on its current step's feature $X_t$, decision $D_t$, outcome of interest $Y_t$, and sensitive $S$. 
\end{Asm}
This assumption, as discussed in previous literature, can be attributed to either bureaucratic policies~\citep{liu2018delayed} or changes in individual behavior, in response to recommendations~\citep{karimi2021causalrecourse} or social learning~\citep{heidari2019effort}.
In the lending context, it is commonly assumed that the higher the credit score the better. Then the assumption is: individuals approved for a loan ($D=1$) experience a positive score change upon successful repayment ($Y=1$) and a negative score change in case of default ($Y=0$), while individuals rejected for a loan ($D=0$) are assumed to have no score change~\citep{liu2018delayed, creager2020causal, damour2020fairness}. In scenarios where individuals who are not granted a loan ($D=0$) seek recourse, it would be assumed that a negative decision leads to an increase in credit score, to elicit a positive decision change in subsequent time steps \citep{heidari2019effort, karimi2021causalrecourse}. 

For the transition probabilities to be time-homogeneous, we take the following assumptions:
\begin{Asm} \label{asm:dynamicsfunct-fixed}
Dynamics $\dynamicfunctall$ remain fixed over time. 
\end{Asm}
This is a common assumption in the literature~\citep{  zhang2020fair,creager2020causal, damour2020fairness,Hu_Zhang_2022}.
Although real-world data often exhibits temporal changes, we make the simplifying assumption of static dynamics. 
We can treat the dynamics as constant for specific durations.
This is reasonable in situations where changes are based on policies involving bureaucratic adjustments \citep{liu2018delayed} or algorithmic recourse recommendations \citep{karimi2020survey}, and where it is desirable for these policies to remain unchanged or not be retrained at every time step \citep{perdomo2020performative}. 
In practical applications, MDPs with time-varying transition probabilities present challenges, and the literature addresses this through online learning algorithms (e.g., \citep{yu2009online, li2019online}).
\begin{Asm}\label{asm:labelfunct-fixed}
    Label distribution $\labelfunctall$ remains fixed over time.
\end{Asm}
This assumption is widely recognized in the literature~\citep{heidari2019effort, zhang2020fair, creager2020causal, damour2020fairness,  karimi2021causalrecourse, Hu_Zhang_2022}. However, in real-world scenarios, the relationship between input data $X_t$ and the target output $Y_t$ may change over time, resulting in changes in the conditional distribution $\labelfunctall$. This phenomenon is commonly referred to as \emph{concept drift}~\citep{lu2018concept, gama2014concept}.
In the lending scenario, concept drift may arise from changes in individuals' repayment behavior or alterations in the process of generating credit scores based on underlying features like income, assets, etc. 
%

\subsection{Additional Example: Qualifications over Time}

\begin{wrapfigure}{r}{0.35\textwidth}
\vspace{-38pt}
    \centering
    \scalebox{0.8}{
        \input{figures/data-gen-y-x}
    }
    \caption{Data gen. model.
    Time steps (subscript) ${t=\{0, 1, 2\}}$.}
    \label{fig:data-gen-y}
    \vspace{-20pt}
\end{wrapfigure}
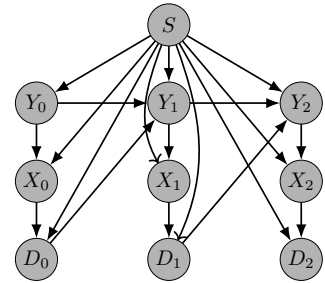
In this section, we provide an additional example, which could also be covered by our framework. The example was provided by~\citep{zhang2020fair} with their data generative model displayed in Figure~\ref{fig:data-gen-y}.
The primary distinction from the example presented in Section~\ref{sec:example} lies in the assumption that $Y_t \rightarrow X_t$. \citep{zhang2020fair} employ their model to replicate lending and recidivism scenarios over time in their experiments, using FICO and COMPAS data, respectively.
However, most prior work has modeled the (FICO) lending examples as $X_t \rightarrow Y_t$~\citep{ liu2018delayed, creager2020causal,damour2020fairness}. The same holds for recidivism (COMPAS)
~\citep{russell2017worlds}. 
We, therefore, frame the example as a repeated admission example where $Y_t$ denotes a (presumably hidden) qualification state at time $t$, following~\citep{rateike2022don, kusner2017counterfactualfairness}. 
\paragraph{Data Generative Model.} Let an individual with protected attribute $S$ (e.g., gender) at time $t$ be described by a qualification $Y_t$ and a non-sensitive feature $X_t$ (e.g., grade or recommendations levels).
We assume the sensitive attribute to remain fixed over time, and drop the attributes time subscript.
For simplicity, we assume binary sensitive attribute and qualification, i.e., ${S, Y_t \in \{0,1\}}$ and a one-dimensional discrete non-sensitive feature $X_t \in \mathbb{Z}$.
Let the population's sensitive attributes be distributed as $\sensitiveall:=\Pbb(S=s)$ and assume them to remain constant over time. We assume $Y_t$ to depend on $S$, such that the group-conditional qualification distribution at time $t$ is $\ystationarytall:=\Pbb(Y_t=y \mid S=s)$.
For example, different demographic groups may have different qualification distributions due to structural discrimination in society. 
We assume that the non-sensitive features $X_t$ are influenced by the qualification $Y_t$ and, possibly (e.g., due to structural discrimination), the sensitive attribute $S$. This leads to the feature distribution $\ylabelfunctall:= \Pbb(X_t=x \mid Y_t=y, S=s)$,
We assume that there exists a policy that takes at each time step $t$ binary decisions $D_t$ (e.g., whether to admit) based on $X_t$ and (potentially) $S$
and decides with probability ${\policyall:=\Pbb(D_t=d\mid X_t=x, S=s)}$. 

Consider now dynamics in which the decision $D_t$ made at one time step $t$, directly impacts an individual's qualifications at the next step, $Y_{t+1}$.  
Assume the transition from the current qualification state $Y_t$ to the next state $Y_{t+1}$ is determined by the current qualification state $Y_t$, decision $D_t$ and (potentially) sensitive attribute $S$.
For example, upon receiving a positive admission decision, an individual may be very motivated and increase their qualifications. However, due to structural discrimination, the extent of the qualification change may be influenced by the individual's sensitive attribute.
We denote the probability of an individual with $S=s$ changing from qualification $Y_t=y$ to $Y_{t+1}=k$ in the next step in response to decision~$D_t=d$ as dynamics ${\ydynamicfunctall:=\Pbb(Y_{t+1} = k\mid Y_{t}=y, D_t=d, S=s)}$.
Crucially, the next step qualification state (conditioned on the sensitive attribute) depends only on the present state qualification and decision, and not on any past states. 

\paragraph{Dynamical System.} 
We can now describe the evolution of the group-conditional qualification distribution $\ystationarytall$ over time $t$. 
The probability of a qualification change from $y$ to $k$ in the next step given $s$ is obtained by marginalizing out decision $D_t$, resulting in 
\begin{align}\label{eq:example-kernel}
     \Pbb(Y_{t+1}=k \mid Y_{t}=y, S=s) = \sum_{xd} \ydynamicfunctall \policyall \ylabelfunctall.
\end{align}
These transition probabilities together with the initial distribution over states $\yinitialdistribution$ define the behavior of the dynamical system.
In our model, we assume that the dynamics $\ydynamicfunctall$ are time-independent, meaning that the qualification changes in response to the decision, the previous qualification and the sensitive attribute remain constant over time.
We also assume that the distribution of the non-sensitive features conditioned on an individual's qualification and sensitive attribute $\ylabelfunctall$ does not change over time (e.g., individuals need a certain qualification to generate certain non-sensitive features). 
Additionally, we assume that the policy $\policyall$ can be chosen by a policy maker and may depend on time. Under these assumptions, the probability of a feature change depends solely on the policy $\policy$ and sensitive feature $S$.

%% file: figures/data-gen-y-x.tex
  \begin{tikzpicture}[every node/.style={inner sep=0,outer sep=0}]

        \node[state, fill=gray!60] (x0) at (0,0) {$Y_0$};
        \node[state, fill=gray!60] (x1) [right  =  1.5 cm of x0]  {$Y_1$};
        \node[state, fill=gray!60] (x2) [right  =  1.5  cm of x1]  {$Y_2$};
        \node[state, fill=gray!60] (s) [above  =  0.6cm of x1]  {$S$};
         \node[state, fill=gray!60] (y0) [below  =  0.6 cm of x0] {$X_0$};
        \node[state, fill=gray!60] (y1) [below  =  0.6 cm of x1]  {$X_1$};
        \node[state, fill=gray!60] (y2) [below  =  0.6 cm of x2]  {$X_2$};
        \node[state,  fill=gray!60] (d0) [below  =  0.6 cm of y0] {$D_0$};
        \node[state,  fill=gray!60] (d1) [below  =  0.6 cm of y1]  {$D_1$};
        \node[state,  fill=gray!60] (d2) [below  =  0.6 cm of y2]  {$D_2$};

        \path (x0) edge [thick](x1);
        \path (x1) edge [thick] (x2);
        
        \path (s) edge [thick](y0);
        \draw [thick, ->] (s) to [out=250,in=120] (y1);
        \path (s) edge [thick](y2);

        \path (s) edge [thick](d0);
        \draw [thick, ->] (s) to [out=290,in=65] (d1);
        \path (s) edge [thick](d2);

        \path (s) edge [thick](x0);
        \path (s) edge [thick](x1);
        \path (s) edge [thick](x2);

        \path (x0) edge [thick](y0);
        \path (x1) edge [thick](y1);
        \path (x2) edge [thick](y2);

        \path  (y0) edge [thick] (d0);
        \path  (y1) edge [thick] (d1);
        \path   (y2) edge [thick] (d2);
        


        \path (d0) edge [thick](x1);
        \path (d1) edge [thick](x2);


\end{tikzpicture}